\documentclass{article}

\usepackage{arxiv}

\usepackage[utf8]{inputenc} %
\usepackage[T1]{fontenc}    %
\usepackage{hyperref}       %
\usepackage{url}            %
\usepackage{booktabs}       %
\usepackage{amsfonts}       %
\usepackage{nicefrac}       %
\usepackage{microtype}      %
\usepackage{cleveref}       %
\usepackage{lipsum}         %
\usepackage{graphicx}
\usepackage{natbib}
\usepackage{doi}

\usepackage[whole]{bxcjkjatype}
\usepackage{graphicx}
\usepackage{cprotect}
\usepackage{booktabs}
\usepackage{longtable}
\usepackage{chngcntr}
\usepackage{color}    %
\usepackage{subfig}

\title{Computational Storytelling and Emotions: A Survey}

\author{Yusuke Mori$^{1}$ \, Hiroaki Yamane$^{2, 1}$ \, Yusuke Mukuta$^{1,2}$ \, Tatsuya Harada$^{1,2}$  \\
         $^{1}$The University of Tokyo,  $^{2}$RIKEN \\
         {\tt \{mori, yamane, mukuta, harada\}@mi.t.u-tokyo.ac.jp}
         }

\renewcommand{\shorttitle}{Computational Storytelling and Emotions: A Survey}

\hypersetup{
pdftitle={Computational Storytelling and Emotions: A Survey},
pdfsubject={},
pdfauthor={Yusuke Mori, Hiroaki Yamane, Yusuke Mukuta, Tatsuya Harada},
pdfkeywords={Computational-Storytelling-and-Emotions-Survey},
}

\begin{document}

\maketitle

\begin{abstract}
Storytelling has always been vital for human nature. From ancient times, humans have used stories for several objectives including entertainment, advertisement, and education. Various analyses have been conducted by researchers and creators to determine the way of producing good stories. The deep relationship between stories and emotions is a prime example. With the advancement in deep learning technology, computers are expected to understand and generate stories. This survey paper is intended to summarize and further contribute to the development of research being conducted on the relationship between stories and emotions.
We believe creativity research is not to replace humans with computers, but to find a way of collaboration between humans and computers to enhance the creativity. With the intention of creating a new intersection between computational storytelling research and human creative writing, we introduced creative techniques used by professional storytellers.
\end{abstract}

\keywords{Storytelling \and Story Completion \and Story Understanding \and Creative Support \and Natural Language Processing}

\section{Introduction}
\label{sec:intro}

What makes a story a story? In this survey, we focus on \textbf{``emotion''} as an important keyword to get to the root of human creativity and to imitate and support it through computers.

The relationship between stories and emotions has been an essential part of research in the field of humanities, representing the cognitive and affective science of literature~\citep{Hogan_2006_10.2307/25115327,Pandit2006,Johnson-Laird_2008-07784-007,Hogan_2010_10.5250/symploke.18.1-2.0065,Hogan2019}.
For creators, the practical knowledge of creative techniques stresses on the importance of being conscious of readers' emotions in order to satisfy them.
These findings have been introduced in computational linguistics (CL), Natural Language Processing (NLP), and Digital Humanities (DH).
For example, in their book ``The Emotion Thesaurus: A Writer's Guide to Character Expression,''~\citet{angela_and_becca_emotion_thesaurus} insisted that a key component of every character is emotion. Referring to this,~\citet{kim-klinger-2019-analysis} analyzed how emotions are expressed non-verbally in a corpus of fan fiction short stories.

\citet{Lugmayr_2017_serious_storytelling}~insisted that a fundamental aspect of storytelling is the emotion, that allows the story to evoke cognitive aspects in its audience. With regard to serious storytelling, they referred to Denning's book~\citep{2005leader}, which stated that having no emotional elements in presentations and applications can cause a failure in project funding.
Therefore, numerous efforts have been made to disclose the mystery of the relationship between emotions and stories~\citep{Anderson_and_McMaster_1982,Strapparava:2008:LIE:1363686.1364052,abdulmageed-ungar:2017:Long,kim-klinger-2018-feels,kim-klinger-2019-frowning,zad-finlayson-2020-systematic}.

\begin{figure}[!t]
    \begin{center}
    \includegraphics[clip,width=0.8\textwidth]{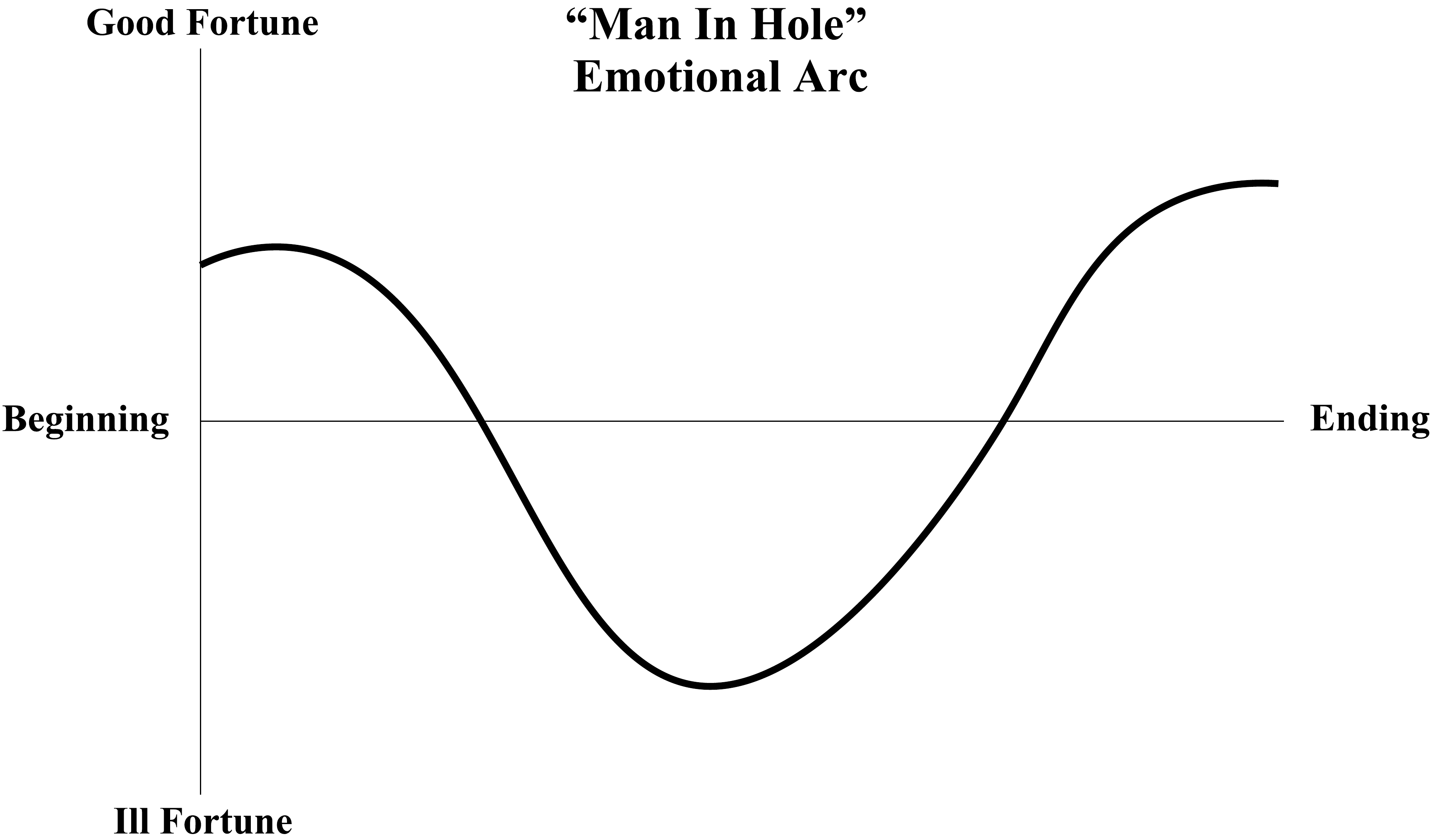}
    \end{center}
    \caption{A representative emotional arc, called ``Man In Hole.'' It starts a little above average. Somebody gets into trouble and gets out of it. This is a reproduction of a diagram drawn by \citet{Vonnegut1981_video}.}
    \label{fig:man-in-hole}
\end{figure}

\citet{MOHAMMAD2012730}~showed that sentiment analysis could be used along with effective visualizations to quantify and track emotions in mails and books.
Referring to the talk by Kurt Vonnegut~\citep{Vonnegut1981_video}, a famous American writer, attempts were made to classify stories by drawing their ``emotional arcs''~\citep{EmotionalArcs,Chu_and_Roy_2017_ICDM,somasundaran-etal-2020-emotion,Vecchio_2020_Hollywood_emotional_arc}. 
In Figure~\ref{fig:man-in-hole}, we show an example of emotional arc.

Characters and their relationship are considered as essential for literary analysis~\citep{bamman-underwood-smith:2014:P14-1,vala-etal-2015-mr,iyyer-EtAl:2016:N16-1,Chaturvedi_AAAI1714564}.
In our previous work~\citep{MORI20191865}, we investigated whether the emotional flow of a story is useful in predicting the reader's interest. 
Moreover, there are some works that have tried to control story generation by considering emotions~\citep{chandu-etal-2019-way,luo-etal-2019-learning,brahman2020modeling,Dathathri2020Plug,MEGATRON-CNTRL}.

There are various terms which are interchangeably used with ``emotion,'' such as sentiment, feeling, and affect.
Furthermore, the field of NLP uses ``sentiment analysis'' and ``emotion analysis'' differently.
Referring to~\citep{Mayer2008,liu_2015}, \citet{Kim2019b}~defined emotion and sentiment as follows:

\begin{itemize}
    \item \textbf{Emotion:} an integrated feeling state involving physiological changes, motor-preparedness, cognition about action, and inner experiences that emerges from an appraisal of the self or situation.
    \item \textbf{Sentiment:} positive or negative feeling underlying an opinion.
\end{itemize}

This definition is followed in this survey, but for simplicity, and unless otherwise noted, the two terms are used interchangeably as follows; ``sentiment'' refers to the positive-negative emotion axis (i.e., 1-dimensional expression), while ``emotion'' refers to the more diverse range of emotions.

In this paper, we discuss the results of our survey, conducted to verify the relationship between stories and emotions. 
Here, we introduce some remarkable recent survey papers on related topics.

\citet{Alhussain_and_Azmi_2021}~surveyed on automatic story generation, especially focusing on non-interactive textual stories. They introduced three groups of such models: the structural models, planning-based models, and machine learning models. In their survey, they introduced MEXICA~\citep{Perez_2001_MEXICA}, MINSTREL~\citep{Turner_1993,turner1994creative}, and the analysis of emotional flow~\citep{MORI20191865} as the stream of understanding and generating stories that incorporate emotions. 

\citet{Bae_2021_preliminary_survey}~conducted a preliminary survey on story interestingness, setting ``how to measure story interestingness'' as the research question. They insisted that there are two factors involved: the cognitive factor and the emotional factor. 

\citet{Li2022_arxiv_survey_text_generation}~focused on the paradigm of pretrained language models (PLMs) and its application in text generation. They discussed the key factors for getting a desirable output.

Our survey is novel because in terms of creative writing support we include the findings of professional creative writers. Furthermore, computational story understanding and generation is an interdisciplinary field between humanities and the information sciences. Although this research is mainly based in the field of information science, it has a viewpoint associated with applications in humanities and practical knowledge to this field.
Recent findings in neuroscience that examine the relationship between emotion and story is also presented. Although emotions are handled in various ways in story comprehension and generation, a more appropriate way to handle emotions in light of the latest findings is also presented in this study.

\begin{figure}[!t]
    \begin{center}
    \includegraphics[clip,width=0.5\textwidth]{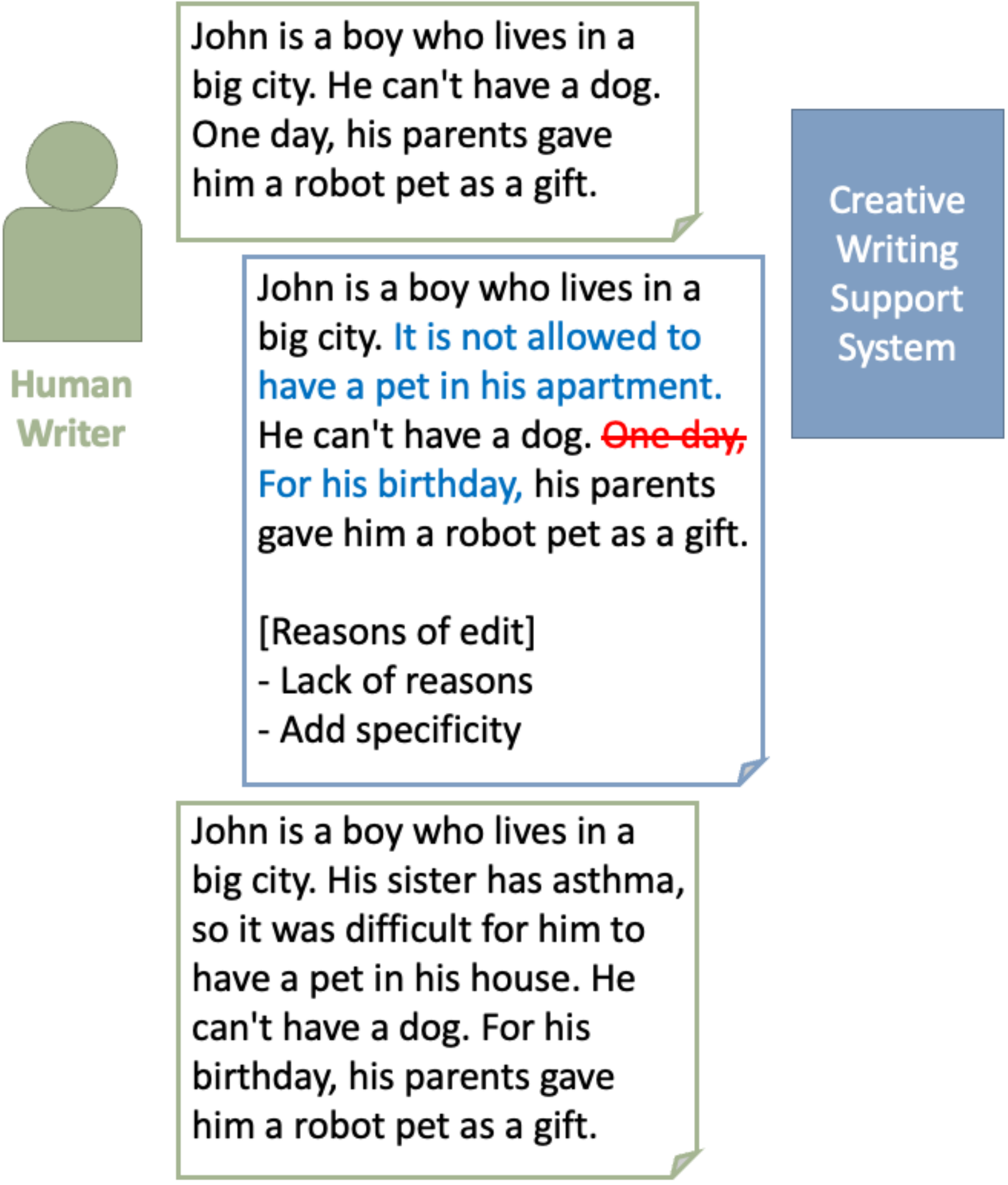}
    \end{center}
    \caption{Schematic diagram of story generation through human-computer interaction.}
    \label{fig:diagram}
\end{figure}

Figure~\ref{fig:diagram} shows a schematic example of human-computer interactive story generation. In this system, the computer edits the text written by the human and then the human re-edits it to create the final version. To achieve this, computers are required to have various abilities such as understanding context, generating appropriate sentences, and explaining why the editing is done.

Now, our research question is; how can computers understand human creativity and enhance it with machine learning technologies?

The main contributions of our survey are as follows:

\begin{itemize}
    \item We conducted our survey on the various elements of a story with particular attention to emotion. With the intention of creating a new intersection between computational storytelling research and human creative writing, we introduced creative techniques used by professional storytellers and examined how emotions are treated in these techniques.
    \item For a computer to support human creativity, it is necessary to solve the integrated task of understanding the story that a human is trying to write and supplementing it with appropriate story generation. This survey was conducted with the intention to accelerate the study on supporting creative writing based on research on story understanding and generation. 
\end{itemize}

Note that our main focus is to understand human creativity with computers and to enhance it with machine learning technologies. We believe creativity research is not about replacing humans with computers but about finding a way in which humans and computers can collaborate to enhance creativity. We hope this survey on stories and emotions will serve as a material to proceed such research.

\section{Structure of this Paper}

As a guide to look through our survey, we briefly describe the significance of each section.

In Section~\ref{sec:emotion_in_screenwriting} we introduce how professional storytellers consider the importance of emotions. 
For concrete examples, we cite some screenwriting techniques and show the relationship between story structure and emotions.

Section~\ref{sec:story_understanding_and_emotions} introduces the research on the relationship between stories and emotions since the 1980s, mainly focusing on three popular views of emotions in computational analysis.

In Section~\ref{sec:story_generation}, we outline recent rapid developments in language models. We present research on story generation; and also discuss text generation. This is to characterize story generation research, which can serve as a reference for future application in other domains to stories. 

In Section~\ref{sec:creative_support}, we introduce the tasks and methods useful in creative support. From the viewpoint of human story writing assistance, we believe in the possibility of a story completion approach. 

Furthermore, we widen our survey to affective neuroscience in Section~\ref{sec:emotions_in_cutting-edge_neuroscience}.
We focus on emotions through the viewpoint of recent advances in affective neuroscience, to introduce an alternative perspective to the three popular views of emotions.

In this paper, we focus on emotion as an important element in a story; however, emotion alone does not make a story. Thus, in Section~\ref{sec:other_aspects_and_future_direction}, we address other elements in a story and discuss their integration with research on emotion and their future direction.

Finally, we conclude our survey in Section~\ref{sec:conclusion}.

\section{Emotion in Screenwriting Theories}
\label{sec:emotion_in_screenwriting}

In research on the creation/co-creation of stories, the findings of those who actually create stories are helpful.
A representative example is ``emotional arcs.''
This term is well-known to have been introduced by the American writer, \citet{Vonnegut1981_video}. 
Based on the idea, \citet{EmotionalArcs}~studied how stories could be clustered.
In this section, we survey some practical creative writing techniques which we believe, will be helpful in conducting research on creative assistance.

Theories about creating stories emphasize the importance of being conscious of readers' emotions in order to satisfy them. 
In particular, theories of screenplay writing are often considered to be practical, even by novelists~\citep{brody2018save}.
This can be attributed to the fact that films require more people and money than novels, and have a restriction on how long a movie can run. 
Thus, there is a stronger need and merit to theorize how to ``construct a structure of stories that sells well'' than novels.

Our target is a novel, a story written in a text.
We are aware that creating a novel is not the same as creating a film script. However, film screenplay writing has indeed been introduced into the practice of novelists and has proven useful in these different media. As a specific example, ``SAVE THE CAT!''~\citep{savethecat}, a technique originally proposed for screenplay writing, has been systematically applied to novel writing~\citep{brody2018save}.\footnote{Tetsuro Shimauchi, who translated~\citep{Karl_2005_writing_for_emotional_impact}~and~\citep{brody2018save}~to Japanese, noted in his afterword as the translator in~\citep{brody2018save} that the response from those who write novels has been more noticeable than those who write screenplays, even though~\citep{Karl_2005_writing_for_emotional_impact} is a book on screenwriting.}

\begin{figure}[!t]
    \begin{center}
    \includegraphics[clip,width=0.9\textwidth]{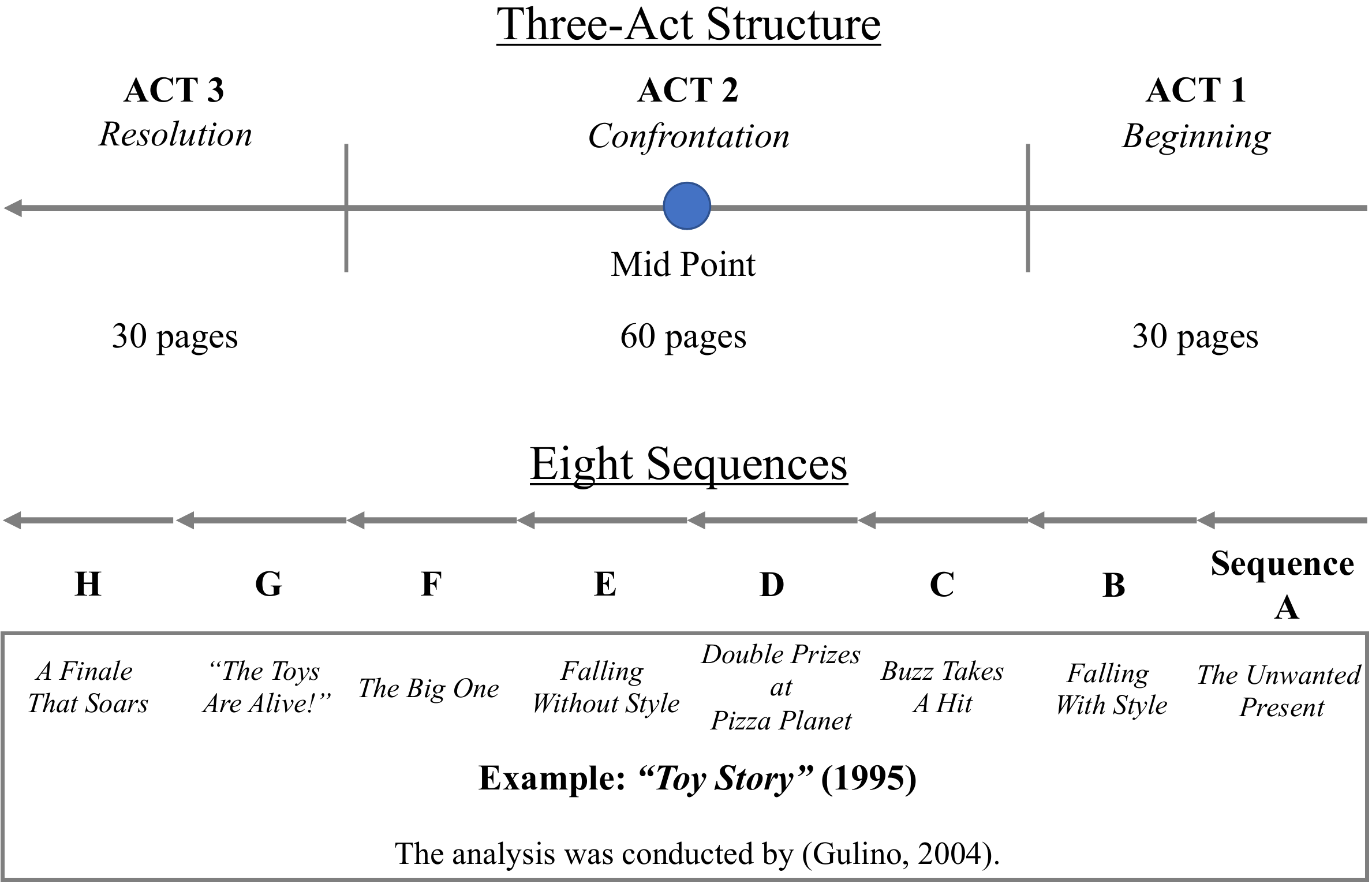}
    \end{center}
    \caption{The overview of the Three-Act Structure and Eight Sequences. As an example of the latter, we used the analysis of ``Toy Story'' (1995) conducted by~\citet{gulino2004screenwriting}.}
    \label{fig:screenwriting}
\end{figure}

\begin{table}[!t]
    \caption{The overview of the Blake Snyder Beat Sheet.}
    \label{tab:bs2}
    \centering
    \begin{tabular}{ccc}
         & \textbf{Beat} & \textbf{Pages} \\
        \hline
        1 & Opening Image & 1 \\
        2 & Theme Stated & 5 \\
        3 & Set-up & 1 - 10 \\
        4 & Catalyst & 12 \\
        5 & Debate & 12 - 25 \\
        6 & Break into Two & 25 \\
        7 & Sub Plot & 30 \\
        8 & Fun and Games & 30 - 55 \\
        9 & Midpoint & 55 \\
        10 & Bad Guys Close In & 55 - 75 \\
        11 & All Is Lost & 75 \\
        12 & Dark Night of the Soul & 75 - 85 \\
        13 & Break into Three & 85 \\
        14 & Finale & 85 - 110 \\
        15 & Final Image & 110 \\
        \hline \\
    \end{tabular}
\end{table}

One of the most famous screenplay structures is the ``three-act structure'' proposed by~\citet{SydField1984}. The first act sets up the situation, the second depicts the conflicts of the characters, and the third depicts the resolution. The entire screenplay is estimated to be 120 pages distributed as 30, 60, and 30 pages. The second act is particularly long; therefore, it is divided into the first and second halves with a central point.
Figure \ref{fig:screenwriting}~shows the overview of the three-act structure.

According to~\citet{gulino2004screenwriting,Gulino2019_video}, ``eight sequences'' were proposed by Frank Daniel. While teaching screenwriting to students based on the three-act structure, he advised the students who were worried that each act was too long for the screenplay, to divide it into eight sequences of 15 pages each. 
Figure \ref{fig:screenwriting}shows the overview of this approach.
This approach divides a story into eight equal lengths.
\citet{gulino2004screenwriting}~explained that another reason for the number of sequences being eight was the physical limitation. 
When movies were shown on films, a 120-minute movie was divided into eight reels (rolls of film) due to capacity constraints. At 18 frames per second, the reels could contain between 10 and 15 minutes. The audience had to wait during the change of reel from the finished one to the next one by the projectionist. 
Filmmakers had to deal with this interval. Each reel needed its role in the movie, therefore, a 120-minute movie came to have eight meaningful units.
Conversely, it can be said that physical constraint was one of the causes that shaped the composition of the film.

Another well-known theory about story structure is ``Blake Snyder Beat Sheet'' (BS2), 
introduced by ``SAVE THE CAT!'' by~\citet{savethecat}.
While the author appreciated that Field's book is helpful for screenwriters, he proposed a new structure with points in between, as the space between acts is too vast for a three-act structure.
This is the division of a story into 15 ``beats,'' and is widely used.
The overview of the BS2 is shown in Table \ref{tab:bs2}.

\citet{Karl_2005_writing_for_emotional_impact} insisted the importance to distinguish between two types of emotions -- character emotions and reader emotions. 
He gave the example of a comedy, where characters are stressed, but the viewers laugh at them. In a thriller, the characters may be calm and unaware, while the viewers may be on the edge of their seats because of a threat that the character is not aware of. He warned that screenwriters, who have a general inkling that emotions are important in a screenplay, put too much attention on the emotions of their characters.
He wrote, \textit{``Whether your character cries or not is not as important as whether the reader cries.''}

\section{Story Understanding and Emotions}
\label{sec:story_understanding_and_emotions}

In this section, we introduce the development of research in the field of story understanding, particularly in terms of understanding emotions. 
We also introduce datasets that handle the relationship between text and emotion and how they have been constructed with different intentions.

\subsection{Understanding Emotion in Stories and Narratives}
\label{subsection:emotions_in_stories}

\begin{table*}[!t]
  \caption{The three popular viewpoints of emotions used in emotion analysis of text.}
  \label{tab:emotion_theory}
  \center
  \begin{tabular}{p{0.25\textwidth}p{0.15\textwidth}p{0.55\textwidth}}
	\textbf{Theory} & \textbf{Type} & \textbf{Features} \\
	\hline
    \textbf{Theory of basic emotions} by~\citet{Ekman1993} & Categorical & Emotions are universal and innate because they are perceived through facial expressions in the same way regardless of what culture one belongs to. There are six emotions: anger, disgust, fear, happiness, sadness, and surprise. \\
    \hline
    \textbf{Wheel of emotions} by~\citet{plutchik1980emotion} & Categorical & A hybrid model that arranges emotions into concentric circles with the inner being the basic and the outer being more complex emotions. \\
    \hline
    \textbf{Circumplex model of affect} by~\citet{Russell1980} & Dimensions & Emotions are represented as vectors in a two-dimensional space and can be explained by a combination of two independent axes, \textit{Valence} and \textit{Arousal}. Valence means pleasantness–unpleasantness and Arousal means calmness–excitement. They are called ``core affect''. As an extension, a third axis called Dominance can be added for a three-dimensional representation~\citep{RUSSELL1977273}. Dominance means perceived control ranging from submissive to dominant. \\
    \hline \\
  \end{tabular}
\end{table*}

A story consists of three elements -- character, event, and setting (i.e., location and temporal setting)~\citep{Park_2020_BigComp}. 
At the intersection of NLP and literary analysis, various studies have been conducted on these factors.
Some analyze stories from the importance of characters and their relationships~\citep{bamman-oconnor-smith:2013:ACL2013,bamman-underwood-smith:2014:P14-1,Massey_2015_DBLP:journals/corr/MasseyXBS15,vala-etal-2015-mr,Chaturvedi_2016_AAAI1612408,Srivastava_2016_AAAI1612173}.
To better comprehend stories, some studies considered stories to be collections of events. 
Using film scripts as a departure point for more general narrative research, \citet{MURTAGH2009302} analyzed their style and structure. They quantified various central perspectives, which have been discussed in McKee's book, ``Story''~\citep{McKee_1997_Story}.

Referring to The Narrative Cloze Test~\citep{NarrativeClozeTest} as a typical example of story understanding task considering events, \citet{mostafazadeh2016} proposed the Story Cloze Test (SCT) as a more difficult task. SCT presents four sentences and the last sentence is excluded from a story composed of five sentences. The system must select an appropriate sentence from two choices that complement the missing last sentence.
In addition to the task, the authors released a large-scale story corpus named ROCStories,
which is a collection of non-fictional daily-life stories written by hundreds of workers of Amazon Mechanical Turk. The five-sentence stories contained varied common-sense knowledge.

\citet{Park_2020_BigComp} analyzed emotions in a story text using an emotion embedding model.
In recent years, emotion estimation has been actively studied~\citep{Park_2019_arXiv,demszky-etal-2020-goemotions} using unsupervised pre-trained large neural models such as BERT~\citep{devlin-etal-2019-bert}.

The number of characters in a story is not always one (protagonist) and research has been done in relation to this.
\citet{Winston_2011_AAAI_Hypotheses} arranged for the simultaneous reading of stories by two separate personas, Dr. Jekyll and Mr. Hyde. 
Presenting NovelPerspective, a tool to subset digital literature based on the point of view (POV), \citet{white-etal-2018-novelperspective} showed many novels have multiple characters, each with their own storyline.
We can refer to them and say that various perspectives should be considered when we consider emotions.

The importance of emotions in storytelling has long been pointed out and efforts to verify it using computers can be traced back to the 1980s. The pioneering work was done by~\citet{Anderson_and_McMaster_1982}.\footnote{Alternatively, when \citet{LEHNERT1981293} proposed Plot Units, he mentioned the events as positive or negative; ``Events that please'' and ``Events that displease'', so this may be called the first step.} Based on the 1,000 most frequently used English words for which \citet{Heise1965} had presented semantic differential factor scores, they reported the development of a computer program to assist the analysis and modeling of emotional tone in text by identifying those words in a passage of discourse. Emotion analysis in text is closely related to lexicons annotated with emotions, and in these annotations and analyses, psychological findings have been referred.

\citet{Kim2019b} stated that there are three theories of emotions popular in computational analysis of emotions: Ekman's theory of basic emotions~\citep{Ekman1993}, Plutchik's wheel of emotion~\citep{plutchik1980emotion}, and Russell's circumplex model~\citep{Russell1980}.\footnote{Russell noted in his paper that: ``While this article was under editorial review, Plutchik (1980) published a specific circumplex model of affect and reported previously unpublished data from H. R. Conte's (1975) doctoral dissertation supporting the model. Their model is similar, but not identical, to the one proposed here.'' It is so impressive that the theories representing categorical emotions and those representing dimensional emotions were proposed at about the same time.}

These three views can be divided into categorical and dimensional emotions \citep{buechel-hahn-2017-emobank,Hakak_2017_survey}: Categorical emotions are represented by Ekman's basic emotions and Plutchik's wheel of emotions. Dimensional emotions are represented by Russell's circumplex model. 
Categorical emotion theories divide emotions into discrete emotion labels.
However, the dimensional emotion theory represents the emotion classes in a dimensional form.
We summarize the three theories of emotions in Table~\ref{tab:emotion_theory}.

Research on emotion and text strives to better understand human-written texts~\citep{Strapparava:2008:LIE:1363686.1364052,abdulmageed-ungar:2017:Long}.
In the domain of story or narrative, \citet{emotions-from-text} provided children's novels with positive/negative sentiment evaluations and a more complex set of eight classes of emotions based on Ekman's basic emotions~\citep{Ekman1993}. To predict emotions using machine learning, they classified the eight emotions again into three emotional valences: positive, negative, and neutral.
\citet{chaturvedi-peng-roth:2017:EMNLP2017} proved that by considering emotional movement in a story, models could improve their performance on SCT. Studies have previously investigated the relationship between emotions and stories' interestingness.
As aforementioned in Section~\ref{sec:emotion_in_screenwriting}, \citet{EmotionalArcs} showed that stories collected from Project Gutenberg\footnote{\url{https://www.gutenberg.org}} could be classified into six styles by considering their emotional arcs (i.e., the trajectory of average happiness in a story).

It has been pointed out that the emotions of a story need to be considered from different perspectives~\citep{Bailey1999SearchingFS,Winston_2011_AAAI_Hypotheses,MORI20191865}. Emotions differ depending on the position of the characters, though the emotions of the reader and the characters may not necessarily coincide. For example, the protagonist and the antagonist may have conflicting emotions. When the protagonist is in a pinch and depressed, the reader may be excited in anticipation of a subsequent counterattack.

In addition, \citet{Winston_2011_AAAI_Hypotheses}, who posited the Strong Story Hypothesis, examined that the interpretation of the story changes depending on the perspective of each reader.
Besides through the Strong Story Hypothesis, he also posited the \textit{Directed Perception Hypothesis}: the mechanisms that enable humans to direct the resources of their perceptual systems to answer questions about real and imagined events account for much of common sense knowledge.
Based on these hypotheses, he and his students---the Genesis Group---proposed Genesis, a story understanding system. They used Shakespeare's ``Macbeth'' as an example and showed that the Genesis could read patterns such as ``vengeance.''
They arranged for the simultaneous reading of stories by two separate personas, which they jocularly named Dr. Jekyll and Mr. Hyde. This showed that the meaning of events in the story changes depending on the perspective.
Finally, he insisted that the two hypotheses are inseparable and gathered them in another hypothesis, the \textit{Inner Language Hypothesis}: human intelligence is enabled by a symbolic inner language faculty whose mechanisms support both story understanding and the querying of perceptual systems.

We~\citep{MORI20191865} considered not only positivity and negativity but also more complicated emotions, using two annotation axes---valence and arousal---based on Russell's circumplex model~\citep{Russell1980}. 
Extending the emotional arc, which deals with only one dimension (positive-negative), we proposed an emotional flow that expresses the emotions of a story in two dimensions (valence and arousal), and discussed the influence of arousal on the interest of the story.
The relationship between emotions and stories is important, not only for those two but also for variable aspects.
In the context of the story ending generation, we conducted a human evaluation and analyzed the reasons due to which evaluators consider a story good or not \citep{mori-etal-2019-toward}. The analysis indicated that human evaluators were conscious of emotions, morals, and common sense when evaluating story endings.
Even human-written endings were judged as not appropriate by human evaluators when they were against morals.
\citet{Hogan_2020_narrative_emotion_ethics} discussed the strong relationship between ethics and ``emotion and narrative'' from the perspective of cognitive and affective science of literature. 

For reference, we would like to touch on research considering reader's emotions in other text domain like news articles~\citep{Lin_2007_SIGIR_emotion_news,Lin_2008_4740453,chang-etal-2015-linguistic}, microblogs~\citep{tang-chen-2012-mining}.
\citet{Lin_2007_SIGIR_emotion_news} studied the classification of news articles into emotions they invoke in their readers. 
Tang and Chen \citep{tang-chen-2012-mining}~focused on a microblog conversation and insisted that the writer of the microblog and the reader who replies to it can both express their emotions. They named the process of changing from writer-emotion to reader-emotion as ``a writer-reader emotion transition'', and examined the mined sentiment words.
Based on recognizing, reader-emotion is different and maybe more complex than writer-emotion~\citep{Lin_2008_4740453,tang-chen-2012-mining}. \citet{chang-etal-2015-linguistic}~worked on reader-emotion estimation and on writing assistance using template considering reader-emotion in the news article domain.
The research of other domains also influences our research on stories and we hope this study influences them in the future.

\subsection{Datasets of Text and Emotion}
\label{subsection:datasets_of_text_and_emotion}

Many datasets have been proposed to deal with the relationship between text and emotion \citep{de-bruyne-etal-2020-emotional}. There are lexicons that give emotional values to words and others that annotate emotions at the sentence level.

As noted in the previous subsection, models of emotion are commonly subdivided into categorical and dimensional ones.
Datasets of text with emotion annotation are generally based on these emotional theories.
Both the categorical and dimensional text-emotion datasets along with lexicons with emotions are presented below.

\begin{itemize}
  \item Categorical text-emotion datasets: (children stories)~\citep{emotions-from-text}, \textit{DENS}~\citep{liu-etal-2019-dens}, \textit{GoEmotions}~\citep{demszky-etal-2020-goemotions}
  \item Dimensional text-emotion datasets: (Facebook posts)~\citep{preotiuc-pietro-etal-2016-modelling}, \textit{EmoBank}~\citep{buechel-hahn-2017-emobank,buechel-hahn-2017-emobank-readers-vs-writers-vs-texts}, \textit{Shared-Character Stories}\footnote{Note that it is part of this dataset that is annotated with emotions.}~\citep{MORI20191865}
  \item Lexicons: the General Inquirer~\citep{General_Inquirer}, LIWC~\citep{LIWC2001,Tausczik_2010_LIWC,LIWC2015}, ANEW~\citep{bradley1999affective}, SentiWordNet~\citep{esuli2006sentiwordnet,baccianella2010sentiwordnet}, SenticNet~\citep{cambria2010senticnet,cambria2020senticnet}, NRC Hashtag Emotion Lexicon~\citep{mohammad:2012:STARSEM-SEMEVAL,mohammad2015using}, NRC Valence, Arousal, Dominance Lexicon~\citep{mohammad-2018-obtaining}, NRC Affect Intensity Lexicon~\citep{LREC18-AIL}
\end{itemize}

For the three categories in the list above, we will mainly focus here on the first two, which are text datasets.

\citet{emotions-from-text} provided children's novels with positive and negative sentiment evaluations and a more complex set of eight classes of emotions based on Ekman's basic emotions \citep{Ekman1993}.
The goal of their annotation project was set to approximately 185 children's stories, including Grimms', H.C. Andersen's, and B. Potter's stories.\footnote{In their paper, they used a preliminary annotated and tie-broken dataset of 1,580 sentences, or 22 of Grimms' tales.}

\citet{liu-etal-2019-dens} introduced the Dataset for Emotions of Narrative Sequences (DENS) for multi-class emotion analysis from long-form narratives in English. DENS was collected from both classic literature available on Project Gutenberg and modern online narratives available on Wattpad, annotated using Amazon Mechanical Turk.
The dataset was annotated based on a modified Plutchik's wheel of emotions \citep{plutchik1980emotion}.
They conducted an initial survey based on 100 stories with a significant fraction sampled from the romance genre. They asked readers to identify the major emotion exhibited in each story, from a choice of the original eight primary emotions: joy, sadness, anger, fear, anticipation, surprise, trust, and disgust. They found that readers have significant difficulty in identifying \textit{trust} as an emotion associated with romantic stories. They modified the annotation scheme by removing Trust and adding Love. They also added the Neutral category to denote passages that do not exhibit any emotional content. The final annotation categories for DENS were: joy, sadness, anger, fear, anticipation, surprise, love, disgust, and neutral.

\citet{demszky-etal-2020-goemotions} proposed \textit{GoEmotions}, the manually annotated dataset of 58,000 English Reddit comments, labeled for 27 emotion categories or neutral, with comments extracted from popular English subreddits.\footnote{\url{https://github.com/google-research/google-research/tree/master/goemotions}}

\citet{buechel-hahn-2017-emobank} explained that dimensional models consider affective states to be best described relative to a small number of independent emotional dimensions (often two or three): Valence (the degree of pleasantness or unpleasantness of an emotion), Arousal (the degree of calmness or excitement), and Dominance (the degree of perceived control ranging from submissive to dominant).\footnote{They noted that the Dominance dimension is sometimes omitted (as we did in \citep{MORI20191865}).}
Based on this idea, \textit{EmoBank} is proposed as a text corpus, manually annotated with emotion according to the psychological VAD scheme \citep{buechel-hahn-2017-emobank,buechel-hahn-2017-emobank-readers-vs-writers-vs-texts}.\footnote{\url{https://github.com/JULIELab/EmoBank}}
They claimed that the VAD model has a major advantage as the dimensions are designed as independent, so results remain comparable dimension-wise even in the absence of others (e.g., Dominance).
Based on the findings of \citet{katz-etal-2007-swat} that annotations are done in vastly different ways depending on the viewpoint of the annotator, they did a pilot study~\citep{buechel-hahn-2017-emobank-readers-vs-writers-vs-texts} on two samples (movie reviews and a genre-balanced corpus) to compare the inter-annotator agreement (IAA) resulting from different annotation perspectives---the writer's and the reader's perspective---in different domains.
They had crowd workers annotate each of the 10,548 sentences from seven domains: news headlines, blogs, essays, fiction, letters, newspapers, and travel guides.
They used the modified 5-point self-assessment manikin (SAM) scales for valence, arousal, and dominance. The original SAM was a 9-point scale~\citep{SAM1994}, but they changed the number to reduce the cognitive load during decision-making for crowd workers. 
From a psychological view, they filtered out highly improbable ratings, which were considered as fraudulent responses. Finally, they got a total of 10,062 sentences bearing VAD values for both perspectives.

We proposed \textit{Shared-Character Stories} dataset (759 stories) and annotations for its subset (100 stories and 623 annotations) \citep{MORI20191865}.\footnote{\url{https://github.com/mil-tokyo/SharedCharacterStories}}
The annotation was based on the psychological Valence-Arousal (VA) model and considered emotional changes in context. 
We also used SAM, but with the original 9-point scale.
Our annotation procedure included emotions and other aspects---\textbf{Storyness:}\footnote{We defined this as ``Storyness'', as stated in \citep{MORI20191865}, but later an anonymous reviewer of our other paper pointed out that ``Storylike-ness'' would be more appropriate. We have retained the term ``Storyness'' because it is used in the published paper and we want to avoid confusion, but we thank the anonymous reviewer for the suggestion.} Does the text seem to be a story? \textbf{Fluency:} Does the story read smoothly and fluently? \textbf{Consistency:} Is the story coherent from sentence to sentence? \textbf{Clarity:} Is the content of the stories easy to understand? \textbf{Meaning:} Does the story have a meaning/message?

We also asked crowd workers to tell us how interesting a story is and write a review.
This made it possible to analyze the relationship between emotions and interestingness (and other aspects).

It is costly and time-consuming to manually annotate various information in texts. Annotating emotions is difficult because ``emotions'' vary from reader to reader (whether it is the reader's own emotions or the understanding of the characters' emotions), making it difficult to determine what is ``correct emotion.'' 
Here we show the examples by~\citet{katz-etal-2007-swat}: while the headline ``Hundreds killed in earthquake'' would be universally accepted as negative, the headline ``Italy defeats France in World Cup Final,'' can be seen as positive, negative, or even neutral depending on the viewpoint of the reader. They insisted that these types of problems made it very difficult for their annotators to provide consistent labels.
Therefore, datasets of text and emotion are much smaller than tasks such as translation and summarization.

Regarding the relationships between categorical and dimensional emotions, 
\citet{LREC18-AIL} stated that joy words have lower arousal scores on an average than sad words, which further have lower average arousal scores than anger and fear. Anger and fear have a very similar profile of average VAD scores. 

Although the traditional basic emotions are an important way of thinking about emotions, the latest affective neuroscience findings deny that everyone has the same emotions regardless of culture or language. This will be discussed in detail in the Section \ref{sec:emotions_in_cutting-edge_neuroscience}. We believe that dimensional emotions can be used to handle a wider variety of emotions and we also emphasize the importance of ``who is feeling.'' In particular, in the case of stories, the reader's emotions are important.

\section{Recent Progress in Computational Storytelling}
\label{sec:story_generation}

Although there are still issues to be considered such as coherence in long text generation, a transformer-based approach has shown to be able to generate sentences that look as if they were written by humans \citep{Otter2020_survey}. For example, \citet{bena-kalita-2019-introducing} confirmed that GPT-2 can generate high-quality poetry and stimulate the reader's emotions.

The importance of emotions in stories is not underestimated and story generation with sentiment control (positive/negative) has been exceptionally well studied \citep{luo-etal-2019-learning,Dathathri2020Plug,MEGATRON-CNTRL}. 
However, unlike the field of dialogue generation, where many studies tackled various emotion control \citep{ghosh-etal-2017-affect,Zhou_2018_AAAI_EmotionalChattingMachine,MA202050,PENG2020105319,song-etal-2019-generating}, story generation with more variety of emotions was not addressed, up until quite recently.
The control of emotions in story generation has focused on how to control positive-negative emotions. Moreover, categorical or dimensional emotions have not been well considered until recently, as per the best of our knowledge.
\citet{brahman2020modeling} did the first work on emotion-aware storytelling, which considers the emotional arc of the protagonist.

In this section, we introduce the brief history of story generation. Then, we introduce controllable story generation, which is essential for emotion-aware control of storytelling in Subsection. 
Finally, we discuss the works that tackled the task of evaluating story-like texts.  

\subsection{Research on Story Generation}
\label{subsec:story_generation}

As we have discussed in the previous section, there is a significant relationship between story understanding (literary analysis) and emotions.
Then, what about story generation?

Klein's ``automatic novel writer''~\citep{Klein1973AutomaticNW}, Meehan's ``TALE-SPIN''~\citep{meehan1976metanovel,Meehan_1977_IJCAI} Lebowitz's ``Universe''~\citep{LEBOWITZ1984171,LEBOWITZ1985483}, and Turner's MINSTREL~\citep{Turner_1993,turner1994creative} are often referred to as early efforts of automatic storytelling.
From the perspective of computational creativity, \citet{Gervas_2009_storytelling_and_creativity} stated that most systems were concerned with telling stories recognized as typical of the particular genre. He insisted the need to adopt evaluation practices introducing some measurement of novelty and quality.

\citet{mcintyre-lapata-2009-learning} published the pioneering work of the data-driven approach on story generation.
Until then, story generation systems had relied on a large repository of background knowledge containing detailed information about the story plot and its characters. The information was usually handcrafted, but in their paper, McIntyre \& Lapata introduced a data-driven approach.

\citet{Hermann:2015:TMR:2969239.2969428} stated two approaches to solving automatic story generation: the symbolic and the neural (network) approaches.
The above studies can be classified into these two and it seems that the former approach was taken at first. However, the latter became stronger due to the progress of deep learning. 

Applying machine learning to human story writing assistance is an approach where interesting works were published in recent years \citep{roemmele_writing_2016,peng2018towards,yao2019plan,goldfarb2019plan}.
Referring to recurrent neural networks (RNN) as a promising machine learning framework for language generation tasks, \citet{roemmele_writing_2016} envisions the task of narrative auto-completion applied to helping an author write a story. \citet{peng2018towards} proposed an analyze-to-generate framework for controllable story generation. They apply two types of generation control: 1) ending valence control (happy or sad ending), 2) storyline keywords.

\citet{Hou_2019_survey_dl_storygeneration} categorized probabilistic story generation models into three categories: \textbf{Theme-Oriented Models}, \textbf{Storyline-Oriented Models}, and \textbf{Human-Machine Interaction-Oriented Models}.
These models are different with respect to the user constraint, that is, the context given by a user to guide the generation probability of words.
Theme-Oriented Models have static user constraint and no human-computer interaction exists. The representative examples of this category were proposed by~\citet{WritingPromptsDataset,xu-etal-2018-skeleton,yao2019plan}.
Storyline-Oriented Models have static user constraint. The constraint contains complete story plots, such as a set of pictures or an abstract description, or a specific story that needs an ending. \citet{wang2018no,Zhao2018plotstoendings,wang_tcvae} proposed representative models.
Human-Machine Interaction-Oriented Models have dynamic user constraints; the constraints vary through human-computer interaction. The models proposed by~\citet{Clark_2018_creative_Writing_machine_in_the_loop,goldfarb2019plan} were considered representatives.

\citet{xu-etal-2018-skeleton} focused on the problem of narrative story generation. As a special kind of story generation~\citep{Li_2013_AAAI}, their problem requires systems to generate a narrative story based on a short description of a scene or an event. They insisted that they eliminate external materials and consider the complete story generation task as~\citet{mcintyre-lapata-2009-learning}.

Hierarchical approaches, which divide story generation into some steps have been widely studied~\citep{fan-etal-2019-strategies,yao2019plan,goldfarb2019plan,goldfarb2020content}.
\citet{fan-etal-2019-strategies} improved the performance of story generation by separating action and entity generation.
Considering ``Bilbo Baggins'' as an example, they insisted on the difficulty of handling characters. They had rare names and were called in different ways: ``he'' and ``hobbit'' may refer to the same entity.
\citet{yao2019plan} proposed a two-step pipeline for open-domain story generation: 1) story planning, which generates a storyline, represented by an ordered list of words, and 2) surface realization, which composes a story based on the storyline. They proposed a hierarchical generation framework named \textit{plan-and-write} that combines storyline planning and surface realization to generate stories from titles.
Based on~\citep{yao2019plan,holtzman-etal-2018-learning}, \citet{goldfarb2019plan} presented a neural narrative generation system named \textit{Plan-and-Revise} in which humans and computers collaborate to generate stories.

In terms of the relationship with emotions, we should note that \citet{Bailey1999SearchingFS} discussed story generation by considering reader emotion.
The paper insisted that the reader's expectations and questions are important for stories.
The proposed cycle of story-generation consisted of four steps: 1) A search-space of possible next segments was generated from the reader's knowledge-base. 2) The effects of each of these possible next segments, in conjunction with the assertions in the story-so-far, on the patterns of expectations and questions derived by the reader, were analyzed. 3) The segment which produced patterns of expectations and questions which best fit the patterns preferred by the abstract narration heuristics was chosen. 4) The chosen segment was asserted as the next segment of the story. The reader's expectations and questions were updated to take account of the new segment. 
This paper proposed such a cycle; however the search-space becomes too large, which is a drawback.
Time has passed since the cycle was proposed. We believe that it could be used more effectively. First, we believe that improvements in the performance of computational resources and the development of search methods have made it possible to search a large space more efficiently. It is also expected that the development of natural language generation methods improves the quality of the candidates generated and improves the segments of a story as a result of the search.

\subsection{Controllable Story Generation with Transformers}
\label{subsec:controllable_story_generation}

The Transformer \citep{Vaswani_2017_transformer} is the basis of today's significant improvement in NLP, and story generation is no exception. 
Here we give an overview of the Transformer and focus on efforts to control generation, which is essential in story generation.

With the advent of the sequence-to-sequence model (Seq2seq), neural networks have become common as a method for generating natural sentences.
Seq2seq was first proposed for machine translation \citep{Sutskever:2014:SSL:2969033.2969173}; however, it has been widely applied to other tasks in NLP \citep{Vinyals2015conversation}.
There is a variant of Seq2seq, which uses convolutional neural networks (CNN) instead of RNNs~\citep{pmlr-v70-gehring17a}.
Then, by replacing the RNNs (or CNNs) in Seq2seq with self-attention, Transformer gets several advantages \citep{tensorflow_transformer_tutorial}: 1) Layer outputs can be calculated in parallel, instead of a series like RNN. 2) Distant items can affect each other's output without passing through many RNN-steps, or convolution layers. 3) It can learn long-range dependencies.  

Unsupervised pre-trained large neural models, such as BERT and GPT-2 \citep{radford2019language}, were proposed using the Transformer architecture, and soon became the mainstream in NLP.
These pre-trained models were divided into two: the Transformer Encoder (bi-directional architecture) and the Transformer Decoder (left-to-right architecture).
In sequence generation, models using left-to-right architecture \citep{radford2019language,Yang_NIPS2019_8812} are more suitable. However, instead of using only one of the Transformer architectures, attempts to create Seq2seq (Encoder-Decoder) models using unsupervised pre-trained large neural models for initializing each of the encoders and the decoders are becoming the new mainstream \citep{lewis-etal-2020-bart,Rothe2020}.
\citet{lewis-etal-2020-bart} proposed BART, which uses BERT as the encoder and GPT-2 as the decoder, and it showed high performance in tasks such as summarization.
\citet{Rothe2020} proposed to use pre-trained checkpoints of BERT, GPT-2, and RoBERTa for initializing a large Transformer-based Seq2seq model.
\citet{2020t5} proposed T5, which uses task-specific prefixes and can solve many text-to-text tasks in one model.
There are also some papers that proposed new variations of Seq2Seq models based on the Transformer~\citep{qi2020prophetnet,roller2020recipes,zhang2020pegasus}.

There are also works for controlling text generation in unsupervised pre-trained large neural models.
CTRL \citep{keskar2019ctrl} is a pioneering work to control particular aspects of text generation with large-scale language models. 
Based on the large-scale language model MEGATRON~\citep{shoeybi2020megatronlm} and knowledge-enhanced story generation~\citep{Guan_2020_TACL}, MEGATRON-CNTRL~\citep{MEGATRON-CNTRL} was proposed.
\citet{rashkin2020plotmachines} proposed the task of outline-conditioned story generation.
The input only provides a rough sketch of the plot, so models need to generate a story by interweaving the key points provided in the outline. 

Inspired by Plug \& Play Generative Networks (PPGN)~\citep{nguyen2017plug} in computer vision, 
\citet{Dathathri2020Plug} proposed Plug-and-Play Language Model (PPLM), an alternative approach for controlled text generation. Their approach uses the attachment models for pre-trained GPT-2 to control the word probability distribution during the word-by-word generation process. 
Optimization is performed ex post facto in the activation space, therefore no re-training or fine-tuning of the core language model is needed.

Following this trend, methods have been presented to control the output by adding modules for output control without modifying the core model: DELOREAN (DEcoding for nonmonotonic LOgical REAsoNing)~\citep{qin-etal-2020-back}, Side-tuning~\citep{Zhang2020_Side-Tuning}, Auxiliary tuning~\citep{zeldes2020technical}, and GeDi~\citep{krause2021gedi}.
\citet{qin-etal-2020-back} solved two tasks by their proposed DELOREAN: ``Abductive Reasoning'' and ``Counterfactual Reasoning.''

Such controllable story generation is important in terms of taking emotions into account. Story generation with sentiment control (positive/negative) has been exceptionally well studied~\citep{luo-etal-2019-learning,Dathathri2020Plug,MEGATRON-CNTRL}. 
Although fine-grained emotion control of story ending generation was tackled by~\citet{luo-etal-2019-learning}, the control of emotions in story generation has been limited to positive-negative emotions. Categorical or dimensional emotions have not been well considered in this area until recently.
\citet{brahman2020modeling} did the pioneering work on emotion-aware storytelling, which considers the emotional arc of the protagonist.
Their study was the first to model the emotional trajectory of the protagonist in neural storytelling. There are significant differences in their study and ours, with respect to task setting and the approach taken. 
We believe that the future direction should consider the reader's emotion.
The ultimate goal of story generation is to generate a story that satisfies readers as readers' emotions are of utmost importance as suggested by~\citet{MORI20191865}. 

As discussed in Subsection \ref{subsection:emotions_in_stories}, the application of transformers in story understanding and generation has attracted attention in recent years.
Transformer architecture's impact is not only for story generation itself, but also for the evaluation of generated stories. 
In the next subsection, we will discuss the recent trend of evaluation metrics.

\subsection{How to Evaluate Story-like Text}
\label{subsection:story_evaluation}

Evaluation of the generated stories (texts) is also a major topic in story (text) generation.
Human evaluation is considered to be the gold standard; however, to evaluate all the texts generated by various models that contain various parameters is impossible. Human evaluation is costly, time-consuming, and dependent on individual abilities. 
Regarding Amazon Mechanical Turk, which is a crowdsourcing platform, \citet{ippolito-etal-2019-unsupervised} pointed out that the evaluation by average workers is unreliable in the task of story infilling. They reported that their inserted honeypot question showed that performance on the question was close to random guessing. 
It was also stated by \citet{august-etal-2020-exploring} that human evaluation schemes tend to ignore the difference of perspectives, authors, and readers. 

Therefore, automatic metrics to evaluate the day-by-day progress of natural language generation is strongly needed.
However, traditional metrics have a poor correlation with human evaluation, so proper evaluation of NLG is difficult~\citep{liu-EtAl:2016:EMNLP20163,novikova-etal-2017-need,chaganty-etal-2018-price,Gatt_and_Krahmer_suvery_NLG,hashimoto-etal-2019-unifying}. 
In their survey of deep learning applied story generation, \citet{Hou_2019_survey_dl_storygeneration} insisted that the inconsistency of datasets and the lack of effective automated evaluation metrics make it difficult to strictly compare the advantages and disadvantages of each model.
Establishing automatic evaluation for various story generation tasks is so difficult that some papers have referred to this as ``future work''~\citep{chandu-etal-2019-way,mori-etal-2019-toward}.

In the field of creative assistance, it would be desirable to have humans participate in interactive evaluation owing to the nature of the task. However, it is impossible in terms of cost and time to have humans evaluate every candidate sentence that is generated. It is important to have an index of ``good stories'' to narrow down the number to a quantity that can be evaluated by humans.

\section{Toward Creative Support}
\label{sec:creative_support}

\citet{Cavazza_2009_emotional_input} stressed the importance of emotions in an interactive storytelling system. At the time they published the paper, natural language processing was a bottleneck hampering the scalability of interactive storytelling systems. Therfore, they introduced their interaction technique based solely on emotional speech recognition. They have concentrated on a small set of five categories (each corresponding to combinations of valence/arousal): Negative Active, Negative Passive, Neutral, Positive Active, and Positive Passive.

Referring to~\citep{Cavazza_2009_emotional_input}, \citet{chandu-etal-2019-way} adapted a personality trait that became crucial to capture and maintain the audience's interest. They stressed that associating the narrative to a personality instigates a sense of empathy and relatedness.
As there were no story datasets in which personality traits were annotated, they used \textit{Image Chat}, a crowdsourced dataset of human-human conversations about an image with a given personality~\citep{shuster-etal-2020-image}.\footnote{Note: \citep{chandu-etal-2019-way} cited the preprint of \citep{shuster-etal-2020-image} on arXiv, 2018.}
An automatic evaluation of whether a story that reflects personality can be generated was left as future work. They also left human evaluation as an issue for the future.

From the viewpoint of human story writing assistance, we believe in the possibility of a story completion approach, wherein machines get an input of an unfinished story and return it in a completed form.
Inspired by SCT, story ending generation (SEG) was designed as a subtask of story generation~\citep{Zhao2018plotstoendings}. 
Given an incomplete story, where the last sentence is excluded from the original five-sentence story, the objective of the task is to automatically generate the last sentence of this given incomplete story.
Furthermore, based on SEG, \citet{wang_tcvae} proposed a story completion task and investigated the problem of generating the missing story plot at any position in an incomplete story. 
If a middle sentence is missing, the task becomes more difficult, as the system must capture the context both before and after the missing sentence.
In addition to this, research on text infilling has been actively conducted in recent years~\citep{ippolito-etal-2019-unsupervised,donahue-etal-2020-enabling,huang-etal-2020-inset,wang2020narrative}, especially about stories. \citet{ippolito-etal-2019-unsupervised} worked on the method to complement the gap between pre and post contexts, which they call story infilling. To understand and generate narrative/story, \citet{wang2020narrative} tackled controlled narrative/story generation where the model is guided to generate coherent narratives with user-specified target endings by interpolation.\footnote{Note that the term ``story completion'' is also used in the area of Psychology \citep{Clarke_2019_doi:10.1080/14780887.2018.1536378}. In this paper, we use the term as its definition in NLP, NLG.}

\citet{Roemmele2021InspirationTO} focused on an ``inspiration through observation'' paradigm for human interaction with generated
text. Emphasizing ``storiability'' as a desirable feature of the result, \citet{Roemmele2021InspirationTO} tackled story infilling task and concluded that automated models can intervene in the writing process without necessarily replacing human effort.

We should cite Abductive Reasoning as a task that is close to Story Completion. 
\citet{Bhagavatula2020Abductive} recently introduced Abductive Reasoning to the area of NLP~\citep{qin-etal-2020-back} and explained, ``Abductive Reasoning has long been considered to be at the core of understanding narratives \citep{HOBBS199369}, reading between the lines~\citep{Norvig1987,Charniak_and_Shimony_1990}, reasoning about everyday situations~\citep{Peirce1937-PEICPO-7,Andersen_1973}, and counterfactual reasoning~\citep{Pearl_2002,Pearl_and_Mackenzie_2018}. Despite the broad recognition of its importance, the study of abductive reasoning in narrative text has very rarely appeared in the NLP literature.'' 
Then, they presented the first study that investigates the viability of language-based abductive reasoning.
Abductive Reasoning is a task that focuses more on causality and common sense. Although it does not necessarily result in a story by itself, the methods and findings in this task can be fed back to Story Completion.

Below are some recent papers that deal with tasks that are close to (or almost equal to) story generation.
\citet{zhu-etal-2020-scriptwriter} proposed an attention-based model called ScriptWriter. It is also in the context of story generation but is closer to dialogue generation with narrative consideration.
\citet{chakrabarty2020simile} focused on figurative language such as a simile to give readers new insights and inspirations. In the paper, they tackled the problem of simile generation. 
\citet{kar2020multiview} tackled story characterization from movie plot synopses and multi-view multi-labeled reviews. This study suggested the importance of considering that the different recipients of the story have their perspectives.

To develop Human-AI co-creation in the domain of story writing, we should not narrow our perspective to just NLP. Such a system has been studied in the collaborative area of NLP and Human Computer Interaction (HCI). 
\citet{Calderwood2020HowNU} explored how professional fiction writers use generative language models during their writing process.
\citet{Osone_2021_BunCho} proposed ``Buncho,'' an AI supported story co-creation system in Japanese. With their system, users can generate titles and synopses from keywords.
\citet{Yuan_2022_wordcraft} sought to learn how people might use large neural language models, such as GPT-3 for creative writing. They proposed Wordcraft, a text editor which makes use of prompting techniques and UX patterns for the collaboration between users and large language models.
We believe that the practical application of systems that are useful to creators will require the collective knowledge of various research areas, as well as knowledge outside of research (knowledge of practitioners).

\section{Emotions in Cutting-edge Neuroscience}
\label{sec:emotions_in_cutting-edge_neuroscience}

As discussed in Subection~\ref{subsection:emotions_in_stories}, there are three popular views of emotions in computational analysis~\citep{Kim2019b}: Ekman's theory of basic emotions, Plutchik's wheel of emotion, and Russell's circumplex model.
However, from the viewpoint of affective neuroscience, there is still an ongoing debate about how emotions should be defined~\citep{Sander2013}.
We should mention here that cutting-edge neuroscientific findings provide us with a new perspective on emotions.

We begin this section by reviewing the classical theories.
Darwin, famous for proposing the theory of evolution, is exclusively known for ``The Origin of Species''~\citep{darwin1859}. He wrote several other books applying the theory. In ``The Origin of Man and the Selection of his Sex''~\citep{darwin1871} and ``On the Facial Expressions of Man and Animals''~\citep{darwin1872}, Darwin stated that the criterion for an accurate assessment of emotion is the facial expression.
During the 1960s, \citet{Tomkins1964} attempted to verify what they defined as basic emotions by linking them to facial expressions. Their experiment, called the basic emotion method, was designed to measure an ability called emotion recognition. Subjects were shown photographs of faces and asked to judge what emotions were indicated by the facial expressions.
Ekman's basic emotions are well known for their use in basic emotion experiments, with subjects from the Fore tribe in Papua New Guinea. It was concluded that emotions are universal and innate because they are perceived through facial expressions in the same way, regardless of what culture one belongs to. The results of this study still form the basis of much of the current research, as shown in Subsection~\ref{subsection:emotions_in_stories}~and~\ref{subsection:datasets_of_text_and_emotion}. 

The theory of basic emotions, established as described above, occupies an important position in affective computing of the text domain, as described in Subsection~\ref{subsection:emotions_in_stories}.
However, in recent years, findings in cutting-edge neuroscience have cast doubt on the universality and innate nature of emotional cognition~\citep{Barrett2017,Barrett2020}. 
Referring to~\citet{Russell1994IsTU}, \citet{Barrett2017} pointed out that Ekman's theory, that is, emotions are universally recognized from facial expressions, has been questioned from a scientific point of view for more than 20 years.
\citet{Barrett2016_theory_of_constructed_emotion} insisted that the last two decades of neuroscience research have brought us to the brink of a paradigm shift in understanding the workings of the brain and setting the stage to revolutionize our understanding of emotions. Based on the brain's structure and function, Barrett proposed the ``theory of constructed emotion,'' which states that emotions should be modeled holistically as whole brain-body phenomena in context.
The author developed an ``emotional granularity'' framework to understand and investigate individual differences in the valence–arousal circumplex model~\citep{Diab2011}. The framework refers to individual differences in distinguishing between emotional states and how valence and arousal are incorporated into representations of emotion.

\citet{Cowen201702247} introduced a conceptual framework to analyze reported emotional states elicited by 2,185 short videos. Across self-reporting methods, they found that videos reliably elicited 27 distinct varieties of reported emotional experience. Further analyses revealed that categorical labels such as ``amusement'' better capture reports of subjective experience than dimensional emotions.
Although there is a limitation in that this is a self-report, the findings may be helpful for the future treatment of emotion in computational linguistics.

In the area of adaptive control, which studies how the brain coordinates cognitive, emotional, and physiological processes to identify problems in the environment and optimize goal-directed behavior, \citet{CLAYSON201944} conducted an event-related brain potential (ERP) study on performance monitoring and found the importance of considering arousal and valence in studies of affective context and cognitive control.

In addition, a problem called priming has been pointed out. Priming is the problem when an experiment is conducted using the basic emotion method, and the subject is given a small number of choices, which becomes a cheat sheet for the subject~\citep{Gendron2012,doi:10.1080/02699931.2013.763769,NELSON201649}. As the number of choices is not large, after a few trials, the subject learns to choose the ``correct emotion'' for the facial expression that is subsequently presented.
This problem should be considered when annotating datasets in CL and NLP.
When annotating text for emotions, if there are not enough options, the same phenomenon of learning the ``correct emotion'' for face images is likely to occur.
Moreover, human evaluation is often taken as a gold standard in evaluating natural language generation models, but further study is needed to find the more appropriate way to instruct the annotators.

\section{Other Aspects of Stories and Suggestions for Future Directions in this Area}
\label{sec:other_aspects_and_future_direction}

In this paper, we focused on emotions, though we understand that emotions are not the only important aspects of stories.
There are many other aspects to be considered: event, common sense, entity, and so on. 
These aspects have deep relationships with emotions, so we will tackle combining the approaches for these aspects with our emotional storytelling system as future work.

Events have been well-studied for story understanding and generation~\citep{NarrativeClozeTest,Martin_AAAI1817046,sims-etal-2019-literary,Tambwekar_ijcai2019-829}, and the relationship between events and emotions have also been paid attention~\citep{rashkin-etal-2018-event2mind}.
Studies on common sense in stories/narratives have a strong relationship with those of events~\citep{mostafazadeh2016,mostafazadeh-EtAl:2017:LSDSem}, and the focus has been broadened to motivations and emotions~\citep{rashkin-etal-2018-modeling,mostafazadeh2020glucose}.
Centering Theory~\citep{grosz-etal-1995-centering} is the base idea of considering entity in text generation. \citet{Barzilay_2008_local_coherence} showed the importance of an entity for managing local-coherence.
\citet{clark-etal-2018-neural} proposed a neural model for text generation that incorporates context via entities.
To support tasks in natural language processing and the computational humanities, \citet{bamman-etal-2019-annotated,sims-etal-2019-literary,bamman-etal-2020-annotated} proposed LitBank,\footnote{\url{https://github.com/dbamman/litbank}} an annotated dataset of 100 different English literary texts annotated for entity categories (person, location, geo-political entity, facility, organization, and vehicle).

In future, it is conceivable to integrate various elements related to story generation, such as characters, emotions, and events, by \textbf{Narratology}.
Narratology, founded on Propp's folktale theory~\citep{Propp1928} as an origin~\citep{Imabuchi_and_Ogata_conference}, has been already shown as worth considering for story generation~\citep{peinado2005creativity,Gervs2006NarrativeM,Imabuchi_and_Ogata_2012_journal}. Moreover, in recent years, \citet{Ogata2019_book} introduced a new concept \textbf{post-narratorogy}, defined as ``computational and cognitive approaches to narratology'' which is closely related to story generation system. 

How to handle ``emotions'' remains a big problem in computational linguistics. 
Which should we use, categorical emotions or dimensional emotions? 
If categorical, how many categories are suitable?
\citet{brahman2020modeling} used five (4 + 1) basic emotions of NRC AIL lexicon. 
They collected open-ended emotion-phrases from COMET \citep{bosselut-etal-2019-comet}, and mapped these phrases to one of the five basic emotions: anger, fear, joy, sadness, and neutral. 
They also tried to directly use more than 500 unique emotional reactions from COMET, but failed because of the few training examples. If so, is ``5'' the best number? 
We think this is debatable. 
Controlling emotions using VAD is more complicated than using the four emotions; however, we believe it will be meaningful because of the potential for more delicate emotional control.

The relationship between the two modes of expression is still an important subject of research.
In his presentation slide in LREC 2018, \citet{mohammad-2018-LREC_slide} noted that the NRC AIL lexicon (i.e., lexicon of four basic emotions) is useful for studying the relationships between affect dimensions, especially when used in combination with the NRC VAD Lexicon.

As introduced in Section \ref{sec:emotions_in_cutting-edge_neuroscience}, ``emotion'' is an important subject of research in field of neuroscience, especially in relation to the brain \citep{Cowen201702247}.
How to incorporate cutting-edge knowledge in these fields into the treatment of language will also be important.

Although there are debates as to whether categorical or dimensional representations are better, it can be said that Arousal in dimensional emotion has an important meaning for human cognition and affection.
Arousal expresses the rise and fall of emotions. A high Arousal level indicates that humans are excited by strong emotions, whereas Low Arousal means that the emotional stimulus is low and ineffective. These factors are closely related to humans' raw emotions and feelings of liveliness. We \citep{MORI20191865}~hypothesized that human satisfaction with stories could be affected not only by pleasant and unpleasant feelings but also by high and low arousal, and confirmed it with dataset construction, experiment, and analysis.

\section{Conclusion}
\label{sec:conclusion}

Creativity is vital for humans, and writing and reading stories are essential aspects of creativity.
Understanding how humans write and read stories has a tight relationship with understanding humans itself.

In this paper, we conducted extensive survey on stories and emotions. 
The novelty of this survey is that it included perspectives such as applications to creative writing support, professional techniques in storytelling, and findings in neuroscience. 
Our research question was: how computers can understand human creativity and enhance it with machine learning technologies? We believe creativity research is not to replace humans with computers, but to find a way of collaboration between humans and computers to enhance the creativity. 
We believe that this paper will lead to further development on story writing with the support of information science and technology and contribute to researchers and story writers.

\section{Acknowledgements}
\label{acknowledgements}

The authors would like to thank Ryohei Shimizu for helpful discussions.
This work was partially supported by JST AIP Acceleration Research JPMJCR20U3, Moonshot R\&D Grant Number JPMJPS2011, JSPS KAKENHI Grant Number JP19H01115, and JP20H05556 and Basic Research Grant (Super AI) of Institute for AI and Beyond of the University of Tokyo.

\bibliographystyle{abbrvnat}

\begin{thebibliography}{211}
\providecommand{\natexlab}[1]{#1}
\providecommand{\url}[1]{#1}
\csname url@samestyle\endcsname
\providecommand{\newblock}{\relax}
\providecommand{\bibinfo}[2]{#2}
\providecommand{\BIBentrySTDinterwordspacing}{\spaceskip=0pt\relax}
\providecommand{\BIBentryALTinterwordstretchfactor}{4}
\providecommand{\BIBentryALTinterwordspacing}{\spaceskip=\fontdimen2\font plus
\BIBentryALTinterwordstretchfactor\fontdimen3\font minus
  \fontdimen4\font\relax}
\providecommand{\BIBforeignlanguage}[2]{{%
\expandafter\ifx\csname l@#1\endcsname\relax
\typeout{** WARNING: IEEEtranN.bst: No hyphenation pattern has been}%
\typeout{** loaded for the language `#1'. Using the pattern for}%
\typeout{** the default language instead.}%
\else
\language=\csname l@#1\endcsname
\fi
#2}}
\providecommand{\BIBdecl}{\relax}
\BIBdecl

\bibitem[Hogan(2006)]{Hogan_2006_10.2307/25115327}
\BIBentryALTinterwordspacing
P.~C. Hogan, ``Narrative universals, heroic tragi-comedy, and shakespeare's
  political ambivalence,'' \emph{College Literature}, vol.~33, no.~1, pp.
  34--66, 2006. [Online]. Available: \url{http://www.jstor.org/stable/25115327}
\BIBentrySTDinterwordspacing

\bibitem[Pandit and Hogan(2006)]{Pandit2006}
\BIBentryALTinterwordspacing
L.~Pandit and P.~C. Hogan, ``\BIBforeignlanguage{English}{Introduction: morsels
  and modules: on embodying cognition in shakespeare's plays (1)},''
  \emph{\BIBforeignlanguage{English}{College Literature}}, vol.~33, pp.~1+,
  2020/10/19/ 2006, 1, Article. [Online]. Available:
  \url{https://link.gale.com/apps/doc/A142620223/ITOF?u=unitokyo&sid=ITOF&xid=d8ead2c9}
\BIBentrySTDinterwordspacing

\bibitem[Johnson-Laird and Oatley(2008)]{Johnson-Laird_2008-07784-007}
R.~N. Johnson-Laird and K.~Oatley, \emph{Emotions, music, and literature.},
  ser. Handbook of emotions, 3rd ed.\hskip 1em plus 0.5em minus 0.4em\relax New
  York, NY, US: The Guilford Press, 2008, pp. 102--113.

\bibitem[Hogan(2010)]{Hogan_2010_10.5250/symploke.18.1-2.0065}
\BIBentryALTinterwordspacing
P.~C. Hogan, ``A passion for plot: Prolegomena to affective narratology,''
  \emph{symplokē}, vol.~18, no. 1-2, pp. 65--81, 2010. [Online]. Available:
  \url{http://www.jstor.org/stable/10.5250/symploke.18.1-2.0065}
\BIBentrySTDinterwordspacing

\bibitem[Hogan(2019)]{Hogan2019}
\BIBentryALTinterwordspacing
------, ``\BIBforeignlanguage{English}{Description, explanation, and the
  meanings of "narrative"},'' \emph{\BIBforeignlanguage{English}{Evolutionary
  Studies in Imaginative Culture}}, vol.~3, pp. 45+, 2020/10/19/ 2019, 1, 45,
  Critical essay. [Online]. Available:
  \url{https://link.gale.com/apps/doc/A595569753/AONE?u=unitokyo&sid=AONE&xid=0e1a136b}
\BIBentrySTDinterwordspacing

\bibitem[Ackerman and Puglisi(2012)]{angela_and_becca_emotion_thesaurus}
A.~Ackerman and B.~Puglisi, \emph{The Emotion Thesaurus: A Writer's Guide to
  Character Expression}.\hskip 1em plus 0.5em minus 0.4em\relax JADD
  Publishing, 2012.

\bibitem[Kim and Klinger(2019{\natexlab{a}})]{kim-klinger-2019-analysis}
\BIBentryALTinterwordspacing
E.~Kim and R.~Klinger, ``An analysis of emotion communication channels in
  fan-fiction: Towards emotional storytelling,'' in \emph{Proceedings of the
  Second Workshop on Storytelling}.\hskip 1em plus 0.5em minus 0.4em\relax
  Florence, Italy: Association for Computational Linguistics, Aug. 2019, pp.
  56--64. [Online]. Available: \url{https://www.aclweb.org/anthology/W19-3406}
\BIBentrySTDinterwordspacing

\bibitem[Lugmayr et~al.(2017)Lugmayr, Sutinen, Suhonen, Islas~Sedano, Hlavacs,
  and Suero~Montero]{Lugmayr_2017_serious_storytelling}
A.~Lugmayr, E.~Sutinen, J.~Suhonen, C.~Islas~Sedano, H.~Hlavacs, and
  C.~Suero~Montero, ``Serious storytelling - a first definition and review,''
  \emph{Multimedia Tools and Applications}, vol.~76, pp. 15\,707--15\,733, 07
  2017.

\bibitem[Denning(2005)]{2005leader}
S.~Denning, \emph{The Leader's Guide to Storytelling: Mastering the Art and
  Discipline of Business Narrative}, ser. J-B US non-Franchise
  Leadership.\hskip 1em plus 0.5em minus 0.4em\relax John Wiley \& Sons, 2005.

\bibitem[Anderson and McMaster(1982)]{Anderson_and_McMaster_1982}
\BIBentryALTinterwordspacing
C.~W. Anderson and G.~E. McMaster, ``Computer assisted modeling of affective
  tone in written documents,'' \emph{Computers and the Humanities}, vol.~16,
  no.~1, pp. 1--9, Sep 1982. [Online]. Available:
  \url{https://doi.org/10.1007/BF02259727}
\BIBentrySTDinterwordspacing

\bibitem[Strapparava and Mihalcea(2008)]{Strapparava:2008:LIE:1363686.1364052}
\BIBentryALTinterwordspacing
C.~Strapparava and R.~Mihalcea, ``Learning to identify emotions in text,'' in
  \emph{Proceedings of ACMSAC 2008}.\hskip 1em plus 0.5em minus 0.4em\relax New
  York, NY, USA: ACM, 2008, pp. 1556--1560. [Online]. Available:
  \url{http://doi.acm.org/10.1145/1363686.1364052}
\BIBentrySTDinterwordspacing

\bibitem[Abdul-Mageed and Ungar(2017)]{abdulmageed-ungar:2017:Long}
\BIBentryALTinterwordspacing
M.~Abdul-Mageed and L.~Ungar, ``Emonet: Fine-grained emotion detection with
  gated recurrent neural networks,'' in \emph{Proceedings of ACL 2017}.\hskip
  1em plus 0.5em minus 0.4em\relax Vancouver, Canada: Association for
  Computational Linguistics, July 2017, pp. 718--728. [Online]. Available:
  \url{http://aclweb.org/anthology/P17-1067}
\BIBentrySTDinterwordspacing

\bibitem[Kim and Klinger(2018)]{kim-klinger-2018-feels}
\BIBentryALTinterwordspacing
E.~Kim and R.~Klinger, ``Who feels what and why? annotation of a literature
  corpus with semantic roles of emotions,'' in \emph{Proceedings of COLING
  2018}.\hskip 1em plus 0.5em minus 0.4em\relax Santa Fe, New Mexico, USA:
  Association for Computational Linguistics, Aug. 2018, pp. 1345--1359.
  [Online]. Available: \url{https://www.aclweb.org/anthology/C18-1114}
\BIBentrySTDinterwordspacing

\bibitem[Kim and Klinger(2019{\natexlab{b}})]{kim-klinger-2019-frowning}
\BIBentryALTinterwordspacing
------, ``Frowning {F}rodo, wincing {L}eia, and a seriously great friendship:
  Learning to classify emotional relationships of fictional characters,'' in
  \emph{Proceedings of NAACL-HLT 2019}.\hskip 1em plus 0.5em minus 0.4em\relax
  Minneapolis, Minnesota: Association for Computational Linguistics, Jun. 2019,
  pp. 647--653. [Online]. Available:
  \url{https://www.aclweb.org/anthology/N19-1067}
\BIBentrySTDinterwordspacing

\bibitem[Zad and Finlayson(2020)]{zad-finlayson-2020-systematic}
\BIBentryALTinterwordspacing
S.~Zad and M.~Finlayson, ``Systematic evaluation of a framework for
  unsupervised emotion recognition for narrative text,'' in \emph{Proceedings
  of the First Joint Workshop on Narrative Understanding, Storylines, and
  Events}.\hskip 1em plus 0.5em minus 0.4em\relax Online: Association for
  Computational Linguistics, Jul. 2020, pp. 26--37. [Online]. Available:
  \url{https://www.aclweb.org/anthology/2020.nuse-1.4}
\BIBentrySTDinterwordspacing

\bibitem[Vonnegut(1995)]{Vonnegut1981_video}
K.~Vonnegut, ``Kurt vonnegut on the shapes of stories,''
  \url{https://www.youtube.com/watch?v=oP3c1h8v2ZQ}, 1995, video. Accessed:
  October 17, 2020.

\bibitem[Mohammad(2012{\natexlab{a}})]{MOHAMMAD2012730}
\BIBentryALTinterwordspacing
S.~M. Mohammad, ``From once upon a time to happily ever after: Tracking
  emotions in mail and books,'' \emph{Decision Support Systems}, vol.~53,
  no.~4, pp. 730 -- 741, 2012, 1) Computational Approaches to Subjectivity and
  Sentiment Analysis 2) Service Science in Information Systems Research :
  Special Issue on PACIS 2010. [Online]. Available:
  \url{http://www.sciencedirect.com/science/article/pii/S0167923612001418}
\BIBentrySTDinterwordspacing

\bibitem[Reagan et~al.(2016)Reagan, Mitchell, Kiley, Danforth, and
  Dodds]{EmotionalArcs}
\BIBentryALTinterwordspacing
A.~J. Reagan, L.~Mitchell, D.~Kiley, C.~M. Danforth, and P.~S. Dodds, ``The
  emotional arcs of stories are dominated by six basic shapes,'' \emph{EPJ Data
  Science}, vol.~5, no.~1, p.~31, Nov 2016. [Online]. Available:
  \url{https://doi.org/10.1140/epjds/s13688-016-0093-1}
\BIBentrySTDinterwordspacing

\bibitem[{Chu} and {Roy}(2017)]{Chu_and_Roy_2017_ICDM}
E.~{Chu} and D.~{Roy}, ``Audio-visual sentiment analysis for learning emotional
  arcs in movies,'' in \emph{Proceedings of ICDM 2017}, 2017, pp. 829--834.

\bibitem[Somasundaran et~al.(2020)Somasundaran, Chen, and
  Flor]{somasundaran-etal-2020-emotion}
\BIBentryALTinterwordspacing
S.~Somasundaran, X.~Chen, and M.~Flor, ``Emotion arcs of student narratives,''
  in \emph{Proceedings of the First Joint Workshop on Narrative Understanding,
  Storylines, and Events}.\hskip 1em plus 0.5em minus 0.4em\relax Online:
  Association for Computational Linguistics, Jul. 2020, pp. 97--107. [Online].
  Available: \url{https://www.aclweb.org/anthology/2020.nuse-1.12}
\BIBentrySTDinterwordspacing

\bibitem[Del~Vecchio et~al.(2020)Del~Vecchio, Kharlamov, Parry, and
  Pogrebna]{Vecchio_2020_Hollywood_emotional_arc}
M.~Del~Vecchio, A.~Kharlamov, G.~Parry, and G.~Pogrebna, ``Improving
  productivity in hollywood with data science: Using emotional arcs of movies
  to drive product and service innovation in entertainment industries,''
  \emph{Journal of the Operational Research Society}, vol.~72, pp. 1--28, 03
  2020.

\bibitem[Bamman et~al.(2014)Bamman, Underwood, and
  Smith]{bamman-underwood-smith:2014:P14-1}
\BIBentryALTinterwordspacing
D.~Bamman, T.~Underwood, and N.~A. Smith, ``A bayesian mixed effects model of
  literary character,'' in \emph{Proceedings of ACL 2014}.\hskip 1em plus 0.5em
  minus 0.4em\relax Baltimore, Maryland: Association for Computational
  Linguistics, June 2014, pp. 370--379. [Online]. Available:
  \url{http://www.aclweb.org/anthology/P/P14/P14-1035}
\BIBentrySTDinterwordspacing

\bibitem[Vala et~al.(2015)Vala, Jurgens, Piper, and Ruths]{vala-etal-2015-mr}
\BIBentryALTinterwordspacing
H.~Vala, D.~Jurgens, A.~Piper, and D.~Ruths, ``Mr. bennet, his coachman, and
  the archbishop walk into a bar but only one of them gets recognized: On the
  difficulty of detecting characters in literary texts,'' in \emph{Proceedings
  of EMNLP 2015}.\hskip 1em plus 0.5em minus 0.4em\relax Lisbon, Portugal:
  Association for Computational Linguistics, Sep. 2015, pp. 769--774. [Online].
  Available: \url{https://www.aclweb.org/anthology/D15-1088}
\BIBentrySTDinterwordspacing

\bibitem[Iyyer et~al.(2016)Iyyer, Guha, Chaturvedi, Boyd-Graber, and
  Daum\'{e}~III]{iyyer-EtAl:2016:N16-1}
\BIBentryALTinterwordspacing
M.~Iyyer, A.~Guha, S.~Chaturvedi, J.~Boyd-Graber, and H.~Daum\'{e}~III,
  ``Feuding families and former friends: Unsupervised learning for dynamic
  fictional relationships,'' in \emph{Proceedings of NAACL-HLT 2016}.\hskip 1em
  plus 0.5em minus 0.4em\relax San Diego, California: Association for
  Computational Linguistics, June 2016, pp. 1534--1544. [Online]. Available:
  \url{http://www.aclweb.org/anthology/N16-1180}
\BIBentrySTDinterwordspacing

\bibitem[Chaturvedi et~al.(2017{\natexlab{a}})Chaturvedi, Iyyer, and
  III]{Chaturvedi_AAAI1714564}
\BIBentryALTinterwordspacing
S.~Chaturvedi, M.~Iyyer, and H.~D. III, ``Unsupervised learning of evolving
  relationships between literary characters,'' 2017. [Online]. Available:
  \url{https://aaai.org/ocs/index.php/AAAI/AAAI17/paper/view/14564}
\BIBentrySTDinterwordspacing

\bibitem[Mori et~al.(2019{\natexlab{a}})Mori, Yamane, Ushiku, and
  Harada]{MORI20191865}
\BIBentryALTinterwordspacing
Y.~Mori, H.~Yamane, Y.~Ushiku, and T.~Harada, ``How narratives move your mind:
  A corpus of shared-character stories for connecting emotional flow and
  interestingness,'' \emph{Information Processing \& Management}, vol.~56,
  no.~5, pp. 1865 -- 1879, 2019. [Online]. Available:
  \url{http://www.sciencedirect.com/science/article/pii/S0306457318305879}
\BIBentrySTDinterwordspacing

\bibitem[Chandu et~al.(2019)Chandu, Prabhumoye, Salakhutdinov, and
  Black]{chandu-etal-2019-way}
\BIBentryALTinterwordspacing
K.~Chandu, S.~Prabhumoye, R.~Salakhutdinov, and A.~W. Black, ``{``}my way of
  telling a story{''}: Persona based grounded story generation,'' in
  \emph{Proceedings of the Second Workshop on Storytelling}.\hskip 1em plus
  0.5em minus 0.4em\relax Florence, Italy: Association for Computational
  Linguistics, Aug. 2019, pp. 11--21. [Online]. Available:
  \url{https://www.aclweb.org/anthology/W19-3402}
\BIBentrySTDinterwordspacing

\bibitem[Luo et~al.(2019)Luo, Dai, Yang, Liu, Chang, Sui, and
  Sun]{luo-etal-2019-learning}
\BIBentryALTinterwordspacing
F.~Luo, D.~Dai, P.~Yang, T.~Liu, B.~Chang, Z.~Sui, and X.~Sun, ``Learning to
  control the fine-grained sentiment for story ending generation,'' in
  \emph{Proceedings of ACL 2019}.\hskip 1em plus 0.5em minus 0.4em\relax
  Florence, Italy: Association for Computational Linguistics, Jul. 2019, pp.
  6020--6026. [Online]. Available:
  \url{https://www.aclweb.org/anthology/P19-1603}
\BIBentrySTDinterwordspacing

\bibitem[Brahman and Chaturvedi(2020)]{brahman2020modeling}
F.~Brahman and S.~Chaturvedi, ``Modeling protagonist emotions for emotion-aware
  storytelling,'' in \emph{Proceedings of EMNLP 2020}.\hskip 1em plus 0.5em
  minus 0.4em\relax Punta Cana, Dominican Republic: Association for
  Computational Linguistics, November 2020.

\bibitem[Dathathri et~al.(2020)Dathathri, Madotto, Lan, Hung, Frank, Molino,
  Yosinski, and Liu]{Dathathri2020Plug}
\BIBentryALTinterwordspacing
S.~Dathathri, A.~Madotto, J.~Lan, J.~Hung, E.~Frank, P.~Molino, J.~Yosinski,
  and R.~Liu, ``Plug and play language models: A simple approach to controlled
  text generation,'' in \emph{Proceedings of ICLR}, 2020. [Online]. Available:
  \url{https://openreview.net/forum?id=H1edEyBKDS}
\BIBentrySTDinterwordspacing

\bibitem[Xu et~al.(2020)Xu, Patwary, Shoeybi, Puri, Fung, Anandkumar, and
  Catanzaro]{MEGATRON-CNTRL}
P.~Xu, M.~Patwary, M.~Shoeybi, R.~Puri, P.~Fung, A.~Anandkumar, and
  B.~Catanzaro, ``Megatron-cntrl: Controllable story generation with external
  knowledge using large-scale language models,'' in \emph{Proceedings of EMNLP
  2020}.\hskip 1em plus 0.5em minus 0.4em\relax Punta Cana, Dominican Republic:
  Association for Computational Linguistics, November 2020.

\bibitem[Mayer et~al.(2008)Mayer, Roberts, and Barsade]{Mayer2008}
\BIBentryALTinterwordspacing
J.~D. Mayer, R.~D. Roberts, and S.~G. Barsade, ``Human abilities: Emotional
  intelligence,'' \emph{Annual Review of Psychology}, vol.~59, no.~1, pp.
  507--536, 2008, pMID: 17937602. [Online]. Available:
  \url{https://doi.org/10.1146/annurev.psych.59.103006.093646}
\BIBentrySTDinterwordspacing

\bibitem[Liu(2015)]{liu_2015}
B.~Liu, \emph{Sentiment Analysis: Mining Opinions, Sentiments, and
  Emotions}.\hskip 1em plus 0.5em minus 0.4em\relax Cambridge University Press,
  2015.

\bibitem[Kim and Klinger(2019{\natexlab{c}})]{Kim2019b}
\BIBentryALTinterwordspacing
E.~Kim and R.~Klinger, ``A survey on sentiment and emotion analysis for
  computational literary studies,'' \emph{Zeitschrift für digitale
  Geisteswissenschaften}, 2019. [Online]. Available:
  \url{https://zfdg.de/2019_008_v1}
\BIBentrySTDinterwordspacing

\bibitem[Alhussain and Azmi(2021)]{Alhussain_and_Azmi_2021}
\BIBentryALTinterwordspacing
A.~I. Alhussain and A.~M. Azmi, ``Automatic story generation: A survey of
  approaches,'' \emph{ACM Comput. Surv.}, vol.~54, no.~5, May 2021. [Online].
  Available: \url{https://doi.org/10.1145/3453156}
\BIBentrySTDinterwordspacing

\bibitem[P{é}rez~y P{é}rez and Sharples(2001)]{Perez_2001_MEXICA}
R.~P{é}rez~y P{é}rez and M.~Sharples, ``Mexica: A computer model of a
  cognitive account of creative writing,'' \emph{J. Exp. Theor. Artif.
  Intell.}, vol.~13, pp. 119--139, 04 2001.

\bibitem[Turner(1993)]{Turner_1993}
S.~R. Turner, ``Minstrel: A computer model of creativity and storytelling,''
  Ph.D. dissertation, USA, 1993, uMI Order no. GAX93-19933.

\bibitem[Turner(1994)]{turner1994creative}
S.~Turner, \emph{The Creative Process: A Computer Model of Storytelling and
  Creativity}.\hskip 1em plus 0.5em minus 0.4em\relax Lawrence Erlbaum
  Associates, Inc., 1994.

\bibitem[Bae et~al.(2021)Bae, Jang, Kim, and Park]{Bae_2021_preliminary_survey}
B.-C. Bae, S.~Jang, Y.~Kim, and S.~Park, ``A preliminary survey on story
  interestingness: Focusing on cognitive and emotional interest,'' in
  \emph{Interactive Storytelling}, A.~Mitchell and M.~Vosmeer, Eds.\hskip 1em
  plus 0.5em minus 0.4em\relax Cham: Springer International Publishing, 2021,
  pp. 447--453.

\bibitem[Li et~al.(2022)Li, Tang, Zhao, Nie, and
  Wen]{Li2022_arxiv_survey_text_generation}
\BIBentryALTinterwordspacing
J.~Li, T.~Tang, W.~X. Zhao, J.~Nie, and J.~Wen, ``A survey of pretrained
  language models based text generation,'' \emph{CoRR}, vol. abs/2201.05273,
  2022. [Online]. Available: \url{https://arxiv.org/abs/2201.05273}
\BIBentrySTDinterwordspacing

\bibitem[Gulino(2004)]{gulino2004screenwriting}
P.~Gulino, \emph{Screenwriting: The Sequence Approach}, ser. Screenwriting: The
  Sequence Approach.\hskip 1em plus 0.5em minus 0.4em\relax Bloomsbury
  Academic, 2004.

\bibitem[Brody(2018)]{brody2018save}
J.~Brody, \emph{Save the Cat! Writes a Novel: The Last Book On Novel Writing
  You'll Ever Need}.\hskip 1em plus 0.5em minus 0.4em\relax Clarkson Potter/Ten
  Speed, 2018.

\bibitem[Snyder(2005)]{savethecat}
B.~Snyder, \emph{{SAVE THE CAT! The Last Book on Screenwriting You'll Ever
  Need}}.\hskip 1em plus 0.5em minus 0.4em\relax Michael Wiese Productions,
  2005.

\bibitem[Iglesias(2005)]{Karl_2005_writing_for_emotional_impact}
\BIBentryALTinterwordspacing
K.~Iglesias, \emph{Writing for Emotional Impact}.\hskip 1em plus 0.5em minus
  0.4em\relax WingSpan Press, 2005. [Online]. Available:
  \url{https://ci.nii.ac.jp/ncid/BB15467795}
\BIBentrySTDinterwordspacing

\bibitem[Field(2006)]{SydField1984}
S.~Field, \emph{The Screenwriter's Workbook, Revised Edition}.\hskip 1em plus
  0.5em minus 0.4em\relax Delta Trade Paperbacks, 2006.

\bibitem[Gulino(2019)]{Gulino2019_video}
P.~J. Gulino, ``8 sequence approach to writing a screenplay - paul joseph
  gulino,'' \url{https://youtu.be/bLMpNwOIfSY}, 2019, video. Accessed: March
  23, 2021.

\bibitem[Ekman(1993)]{Ekman1993}
P.~Ekman, ``Facial expression and emotion,'' \emph{American Psychologist}, vol.
  48(4), pp. 384--392, 1993.

\bibitem[Plutchik(1980)]{plutchik1980emotion}
R.~Plutchik, \emph{Emotion, a Psychoevolutionary Synthesis}.\hskip 1em plus
  0.5em minus 0.4em\relax Harper \& Row, 1980.

\bibitem[Russell(1980)]{Russell1980}
J.~A. Russell, ``A circumplex model of affect,'' \emph{Journal of personality
  and social psychology}, vol.~39, pp. 1161--1178, 1980.

\bibitem[Russell and Mehrabian(1977)]{RUSSELL1977273}
\BIBentryALTinterwordspacing
J.~A. Russell and A.~Mehrabian, ``Evidence for a three-factor theory of
  emotions,'' \emph{Journal of Research in Personality}, vol.~11, no.~3, pp.
  273 -- 294, 1977. [Online]. Available:
  \url{http://www.sciencedirect.com/science/article/pii/009265667790037X}
\BIBentrySTDinterwordspacing

\bibitem[{Park} et~al.(2020){Park}, {Bae}, and {Cheong}]{Park_2020_BigComp}
S.~{Park}, B.~{Bae}, and Y.~{Cheong}, ``Emotion recognition from text stories
  using an emotion embedding model,'' in \emph{Proceedings of BigComp 2020},
  2020, pp. 579--583.

\bibitem[Bamman et~al.(2013)Bamman, O'Connor, and
  Smith]{bamman-oconnor-smith:2013:ACL2013}
\BIBentryALTinterwordspacing
D.~Bamman, B.~O'Connor, and N.~A. Smith, ``Learning latent personas of film
  characters,'' in \emph{Proceedings of ACL 2013}.\hskip 1em plus 0.5em minus
  0.4em\relax Sofia, Bulgaria: Association for Computational Linguistics,
  August 2013, pp. 352--361. [Online]. Available:
  \url{http://www.aclweb.org/anthology/P13-1035}
\BIBentrySTDinterwordspacing

\bibitem[Massey et~al.(2015)Massey, Xia, Bamman, and
  Smith]{Massey_2015_DBLP:journals/corr/MasseyXBS15}
\BIBentryALTinterwordspacing
P.~Massey, P.~Xia, D.~Bamman, and N.~A. Smith, ``Annotating character
  relationships in literary texts,'' \emph{CoRR}, vol. abs/1512.00728, 2015.
  [Online]. Available: \url{http://arxiv.org/abs/1512.00728}
\BIBentrySTDinterwordspacing

\bibitem[Chaturvedi et~al.(2016)Chaturvedi, Srivastava, III, and
  Dyer]{Chaturvedi_2016_AAAI1612408}
\BIBentryALTinterwordspacing
S.~Chaturvedi, S.~Srivastava, H.~D. III, and C.~Dyer, ``Modeling evolving
  relationships between characters in literary novels,'' 2016. [Online].
  Available:
  \url{https://www.aaai.org/ocs/index.php/AAAI/AAAI16/paper/view/12408}
\BIBentrySTDinterwordspacing

\bibitem[Srivastava et~al.(2016)Srivastava, Chaturvedi, and
  Mitchell]{Srivastava_2016_AAAI1612173}
\BIBentryALTinterwordspacing
S.~Srivastava, S.~Chaturvedi, and T.~Mitchell, ``Inferring interpersonal
  relations in narrative summaries,'' 2016. [Online]. Available:
  \url{https://www.aaai.org/ocs/index.php/AAAI/AAAI16/paper/view/12173}
\BIBentrySTDinterwordspacing

\bibitem[Murtagh et~al.(2009)Murtagh, Ganz, and McKie]{MURTAGH2009302}
\BIBentryALTinterwordspacing
F.~Murtagh, A.~Ganz, and S.~McKie, ``The structure of narrative: The case of
  film scripts,'' \emph{Pattern Recognition}, vol.~42, no.~2, pp. 302 -- 312,
  2009, learning Semantics from Multimedia Content. [Online]. Available:
  \url{http://www.sciencedirect.com/science/article/pii/S0031320308002100}
\BIBentrySTDinterwordspacing

\bibitem[McKee(1997)]{McKee_1997_Story}
R.~McKee, \emph{Story: Substance, Structure, Style and the Principles of
  Screenwriting}.\hskip 1em plus 0.5em minus 0.4em\relax Itbooks, 1997.

\bibitem[Chambers and Jurafsky(2008)]{NarrativeClozeTest}
\BIBentryALTinterwordspacing
N.~Chambers and D.~Jurafsky, ``Unsupervised learning of narrative event
  chains,'' in \emph{Proceedings of ACL-HLT 2008}.\hskip 1em plus 0.5em minus
  0.4em\relax Columbus, Ohio: Association for Computational Linguistics, June
  2008, pp. 789--797. [Online]. Available:
  \url{http://www.aclweb.org/anthology/P/P08/P08-1090}
\BIBentrySTDinterwordspacing

\bibitem[Mostafazadeh et~al.(2016)Mostafazadeh, Chambers, He, Parikh, Batra,
  Vanderwende, Kohli, and Allen]{mostafazadeh2016}
\BIBentryALTinterwordspacing
N.~Mostafazadeh, N.~Chambers, X.~He, D.~Parikh, D.~Batra, L.~Vanderwende,
  P.~Kohli, and J.~Allen, ``A corpus and cloze evaluation for deeper
  understanding of commonsense stories,'' in \emph{Proceedings of NAACL-HLT
  2016}.\hskip 1em plus 0.5em minus 0.4em\relax San Diego, California:
  Association for Computational Linguistics, June 2016, pp. 839--849. [Online].
  Available: \url{http://www.aclweb.org/anthology/N16-1098}
\BIBentrySTDinterwordspacing

\bibitem[Park et~al.(2019)Park, Kim, Jeon, Park, and Oh]{Park_2019_arXiv}
\BIBentryALTinterwordspacing
S.~Park, J.~Kim, J.~Jeon, H.~Park, and A.~Oh, ``Toward dimensional emotion
  detection from categorical emotion annotations.'' \emph{CoRR}, vol.
  abs/1911.02499, 2019. [Online]. Available:
  \url{http://arxiv.org/abs/1911.02499}
\BIBentrySTDinterwordspacing

\bibitem[Demszky et~al.(2020)Demszky, Movshovitz-Attias, Ko, Cowen, Nemade, and
  Ravi]{demszky-etal-2020-goemotions}
\BIBentryALTinterwordspacing
D.~Demszky, D.~Movshovitz-Attias, J.~Ko, A.~Cowen, G.~Nemade, and S.~Ravi,
  ``{G}o{E}motions: A dataset of fine-grained emotions,'' in \emph{Proceedings
  of ACL 2020}.\hskip 1em plus 0.5em minus 0.4em\relax Online: Association for
  Computational Linguistics, Jul. 2020, pp. 4040--4054. [Online]. Available:
  \url{https://www.aclweb.org/anthology/2020.acl-main.372}
\BIBentrySTDinterwordspacing

\bibitem[Devlin et~al.(2019)Devlin, Chang, Lee, and
  Toutanova]{devlin-etal-2019-bert}
\BIBentryALTinterwordspacing
J.~Devlin, M.-W. Chang, K.~Lee, and K.~Toutanova, ``{BERT}: Pre-training of
  deep bidirectional transformers for language understanding,'' in
  \emph{Proceedings of NAACL-HLT 2019}.\hskip 1em plus 0.5em minus 0.4em\relax
  Minneapolis, Minnesota: Association for Computational Linguistics, Jun. 2019,
  pp. 4171--4186. [Online]. Available:
  \url{https://www.aclweb.org/anthology/N19-1423}
\BIBentrySTDinterwordspacing

\bibitem[Winston(2011)]{Winston_2011_AAAI_Hypotheses}
\BIBentryALTinterwordspacing
P.~Winston, ``The strong story hypothesis and the directed perception
  hypothesis,'' 2011. [Online]. Available:
  \url{https://www.aaai.org/ocs/index.php/FSS/FSS11/paper/view/4125}
\BIBentrySTDinterwordspacing

\bibitem[White et~al.(2018)White, Togneri, Liu, and
  Bennamoun]{white-etal-2018-novelperspective}
\BIBentryALTinterwordspacing
L.~White, R.~Togneri, W.~Liu, and M.~Bennamoun, ``{N}ovel{P}erspective:
  Identifying point of view characters,'' in \emph{Proceedings of {ACL} 2018,
  System Demonstrations}.\hskip 1em plus 0.5em minus 0.4em\relax Melbourne,
  Australia: Association for Computational Linguistics, Jul. 2018, pp. 7--12.
  [Online]. Available: \url{https://www.aclweb.org/anthology/P18-4002}
\BIBentrySTDinterwordspacing

\bibitem[Lehnert(1981)]{LEHNERT1981293}
\BIBentryALTinterwordspacing
W.~G. Lehnert, ``Plot units and narrative summarization,'' \emph{Cognitive
  Science}, vol.~5, no.~4, pp. 293 -- 331, 1981. [Online]. Available:
  \url{http://www.sciencedirect.com/science/article/pii/S036402138180016X}
\BIBentrySTDinterwordspacing

\bibitem[Heise(1965)]{Heise1965}
\BIBentryALTinterwordspacing
D.~R. Heise, ``Semantic differential profiles for 1,000 most frequent english
  words.'' \emph{Psychological Monographs: General and Applied}, vol.~79,
  no.~8, pp. 1 -- 31, 1965. [Online]. Available:
  \url{http://search.ebscohost.com/login.aspx?direct=true&db=pdh&AN=2011-19203-001&lang=ja&site=ehost-live}
\BIBentrySTDinterwordspacing

\bibitem[Buechel and Hahn(2017{\natexlab{a}})]{buechel-hahn-2017-emobank}
\BIBentryALTinterwordspacing
S.~Buechel and U.~Hahn, ``{E}mo{B}ank: Studying the impact of annotation
  perspective and representation format on dimensional emotion analysis,'' in
  \emph{Proceedings of EACL 2017}.\hskip 1em plus 0.5em minus 0.4em\relax
  Valencia, Spain: Association for Computational Linguistics, Apr. 2017, pp.
  578--585. [Online]. Available:
  \url{https://www.aclweb.org/anthology/E17-2092}
\BIBentrySTDinterwordspacing

\bibitem[{Hakak} et~al.(2017){Hakak}, {Mohd}, {Kirmani}, and
  {Mohd}]{Hakak_2017_survey}
N.~M. {Hakak}, M.~{Mohd}, M.~{Kirmani}, and M.~{Mohd}, ``Emotion analysis: A
  survey,'' in \emph{2017 International Conference on Computer, Communications
  and Electronics (Comptelix)}, 2017, pp. 397--402.

\bibitem[Alm et~al.(2005)Alm, Roth, and Sproat]{emotions-from-text}
\BIBentryALTinterwordspacing
C.~O. Alm, D.~Roth, and R.~Sproat, ``Emotions from text: Machine learning for
  text-based emotion prediction,'' in \emph{Proceedings of EMNLP-HLT 2005},
  2005. [Online]. Available: \url{http://www.aclweb.org/anthology/H05-1073}
\BIBentrySTDinterwordspacing

\bibitem[Chaturvedi et~al.(2017{\natexlab{b}})Chaturvedi, Peng, and
  Roth]{chaturvedi-peng-roth:2017:EMNLP2017}
\BIBentryALTinterwordspacing
S.~Chaturvedi, H.~Peng, and D.~Roth, ``Story comprehension for predicting what
  happens next,'' in \emph{Proceedings of EMNLP 2017}.\hskip 1em plus 0.5em
  minus 0.4em\relax Copenhagen, Denmark: Association for Computational
  Linguistics, September 2017, pp. 1603--1614. [Online]. Available:
  \url{https://www.aclweb.org/anthology/D17-1168}
\BIBentrySTDinterwordspacing

\bibitem[Bailey(1999)]{Bailey1999SearchingFS}
P.~Bailey, ``Searching for storiness: Story-generation from a reader's
  perspective,'' 1999.

\bibitem[Mori et~al.(2019{\natexlab{b}})Mori, Yamane, Mukuta, and
  Harada]{mori-etal-2019-toward}
\BIBentryALTinterwordspacing
Y.~Mori, H.~Yamane, Y.~Mukuta, and T.~Harada, ``Toward a better story end:
  Collecting human evaluation with reasons,'' in \emph{Proceedings of INLG
  2019}.\hskip 1em plus 0.5em minus 0.4em\relax Tokyo, Japan: Association for
  Computational Linguistics, Oct.{--}Nov. 2019, pp. 383--390. [Online].
  Available: \url{https://www.aclweb.org/anthology/W19-8646}
\BIBentrySTDinterwordspacing

\bibitem[Hogan(2020)]{Hogan_2020_narrative_emotion_ethics}
\BIBentryALTinterwordspacing
P.~C. Hogan, ``{Narrative Universals, Emotion, and Ethics},'' \emph{Poetics
  Today}, vol.~41, no.~2, pp. 187--204, 06 2020. [Online]. Available:
  \url{https://doi.org/10.1215/03335372-8172514}
\BIBentrySTDinterwordspacing

\bibitem[Lin et~al.(2007)Lin, Yang, and Chen]{Lin_2007_SIGIR_emotion_news}
\BIBentryALTinterwordspacing
K.~H.-Y. Lin, C.~Yang, and H.-H. Chen, ``What emotions do news articles trigger
  in their readers?'' in \emph{Proceedings of SIGIR}.\hskip 1em plus 0.5em
  minus 0.4em\relax New York, NY, USA: Association for Computing Machinery,
  2007, p. 733–734. [Online]. Available:
  \url{https://doi.org/10.1145/1277741.1277882}
\BIBentrySTDinterwordspacing

\bibitem[{Lin} et~al.(2008){Lin}, {Yang}, and {Chen}]{Lin_2008_4740453}
K.~H. {Lin}, C.~{Yang}, and H.~{Chen}, ``Emotion classification of online news
  articles from the reader's perspective,'' in \emph{2008 IEEE/WIC/ACM
  International Conference on Web Intelligence and Intelligent Agent
  Technology}, vol.~1, 2008, pp. 220--226.

\bibitem[Chang et~al.(2015)Chang, Chen, Hsieh, Chen, and
  Hsu]{chang-etal-2015-linguistic}
\BIBentryALTinterwordspacing
Y.-C. Chang, C.-C. Chen, Y.-L. Hsieh, C.~C. Chen, and W.-L. Hsu, ``Linguistic
  template extraction for recognizing reader-emotion and emotional resonance
  writing assistance,'' in \emph{Proceedings of ACL-IJCNLP 2015}.\hskip 1em
  plus 0.5em minus 0.4em\relax Beijing, China: Association for Computational
  Linguistics, Jul. 2015, pp. 775--780. [Online]. Available:
  \url{https://www.aclweb.org/anthology/P15-2127}
\BIBentrySTDinterwordspacing

\bibitem[Tang and Chen(2012)]{tang-chen-2012-mining}
\BIBentryALTinterwordspacing
Y.-j. Tang and H.-H. Chen, ``Mining sentiment words from microblogs for
  predicting writer-reader emotion transition,'' in \emph{Proceedings of LREC
  2012}.\hskip 1em plus 0.5em minus 0.4em\relax Istanbul, Turkey: European
  Language Resources Association (ELRA), May 2012, pp. 1226--1229. [Online].
  Available:
  \url{http://www.lrec-conf.org/proceedings/lrec2012/pdf/117_Paper.pdf}
\BIBentrySTDinterwordspacing

\bibitem[De~Bruyne et~al.(2020)De~Bruyne, De~Clercq, and
  Hoste]{de-bruyne-etal-2020-emotional}
\BIBentryALTinterwordspacing
L.~De~Bruyne, O.~De~Clercq, and V.~Hoste, ``\BIBforeignlanguage{English}{An
  emotional mess! deciding on a framework for building a {D}utch
  emotion-annotated corpus},'' in
  \emph{\BIBforeignlanguage{English}{Proceedings of LREC 2020}}.\hskip 1em plus
  0.5em minus 0.4em\relax Marseille, France: European Language Resources
  Association, May 2020, pp. 1643--1651. [Online]. Available:
  \url{https://www.aclweb.org/anthology/2020.lrec-1.204}
\BIBentrySTDinterwordspacing

\bibitem[Liu et~al.(2019)Liu, Osama, and De~Andrade]{liu-etal-2019-dens}
\BIBentryALTinterwordspacing
C.~Liu, M.~Osama, and A.~De~Andrade, ``{DENS}: A dataset for multi-class
  emotion analysis,'' in \emph{Proceedings of EMNLP-IJCNLP 2019}.\hskip 1em
  plus 0.5em minus 0.4em\relax Hong Kong, China: Association for Computational
  Linguistics, Nov. 2019, pp. 6293--6298. [Online]. Available:
  \url{https://www.aclweb.org/anthology/D19-1656}
\BIBentrySTDinterwordspacing

\bibitem[Preo{\c{t}}iuc-Pietro et~al.(2016)Preo{\c{t}}iuc-Pietro, Schwartz,
  Park, Eichstaedt, Kern, Ungar, and
  Shulman]{preotiuc-pietro-etal-2016-modelling}
\BIBentryALTinterwordspacing
D.~Preo{\c{t}}iuc-Pietro, H.~A. Schwartz, G.~Park, J.~Eichstaedt, M.~Kern,
  L.~Ungar, and E.~Shulman, ``Modelling valence and arousal in {F}acebook
  posts,'' in \emph{Proceedings of the 7th Workshop on Computational Approaches
  to Subjectivity, Sentiment and Social Media Analysis}.\hskip 1em plus 0.5em
  minus 0.4em\relax San Diego, California: Association for Computational
  Linguistics, Jun. 2016, pp. 9--15. [Online]. Available:
  \url{https://www.aclweb.org/anthology/W16-0404}
\BIBentrySTDinterwordspacing

\bibitem[Buechel and
  Hahn(2017{\natexlab{b}})]{buechel-hahn-2017-emobank-readers-vs-writers-vs-texts}
S.~Buechel and U.~Hahn, ``Readers vs. writers vs. texts: Coping with different
  perspectives of text understanding in emotion annotation,'' 01 2017.

\bibitem[Stone and Hunt(1963)]{General_Inquirer}
\BIBentryALTinterwordspacing
P.~J. Stone and E.~B. Hunt, ``A computer approach to content analysis: Studies
  using the general inquirer system,'' in \emph{Proceedings of the May 21-23,
  1963, Spring Joint Computer Conference}.\hskip 1em plus 0.5em minus
  0.4em\relax New York, NY, USA: Association for Computing Machinery, 1963, p.
  241–256. [Online]. Available: \url{https://doi.org/10.1145/1461551.1461583}
\BIBentrySTDinterwordspacing

\bibitem[Pennebaker et~al.(2001)Pennebaker, Booth, and Francis]{LIWC2001}
J.~W. Pennebaker, R.~J. Booth, and M.~E. Francis, ``Linguistic inquiry and word
  count: Liwc 2001,'' 2001.

\bibitem[Tausczik and Pennebaker(2010)]{Tausczik_2010_LIWC}
\BIBentryALTinterwordspacing
Y.~R. Tausczik and J.~W. Pennebaker, ``The psychological meaning of words: Liwc
  and computerized text analysis methods,'' \emph{Journal of Language and
  Social Psychology}, vol.~29, no.~1, pp. 24--54, 2010. [Online]. Available:
  \url{https://doi.org/10.1177/0261927X09351676}
\BIBentrySTDinterwordspacing

\bibitem[Pennebaker et~al.()Pennebaker, Booth, Boyd, and Francis]{LIWC2015}
J.~W. Pennebaker, R.~J. Booth, R.~L. Boyd, and M.~E. Francis, ``Linguistic
  inquiry and word count: Liwc2015.''

\bibitem[Bradley and Lang(1999)]{bradley1999affective}
M.~M. Bradley and P.~J. Lang, ``Affective norms for english words (anew):
  Instruction manual and affective ratings,'' Technical report C-1, the center
  for research in psychophysiology~…, Tech. Rep., 1999.

\bibitem[Esuli and Sebastiani()]{esuli2006sentiwordnet}
A.~Esuli and F.~Sebastiani, ``Sentiwordnet: A publicly available lexical
  resource for opinion mining.''\hskip 1em plus 0.5em minus 0.4em\relax
  Citeseer.

\bibitem[Baccianella et~al.(2010)Baccianella, Esuli, and
  Sebastiani]{baccianella2010sentiwordnet}
S.~Baccianella, A.~Esuli, and F.~Sebastiani, ``Sentiwordnet 3.0: an enhanced
  lexical resource for sentiment analysis and opinion mining.'' in \emph{Lrec},
  vol.~10, no. 2010, 2010, pp. 2200--2204.

\bibitem[Cambria et~al.(2010)Cambria, Speer, Havasi, and
  Hussain]{cambria2010senticnet}
E.~Cambria, R.~Speer, C.~Havasi, and A.~Hussain, ``Senticnet: A publicly
  available semantic resource for opinion mining,'' in \emph{AAAI Fall
  Symposium: Commonsense Knowledge}, 2010.

\bibitem[Cambria et~al.(2020)Cambria, Li, Xing, Poria, and
  Kwok]{cambria2020senticnet}
\BIBentryALTinterwordspacing
E.~Cambria, Y.~Li, F.~Z. Xing, S.~Poria, and K.~Kwok, ``Senticnet 6: Ensemble
  application of symbolic and subsymbolic ai for sentiment analysis,'' in
  \emph{Proceedings of CIKM}.\hskip 1em plus 0.5em minus 0.4em\relax New York,
  NY, USA: Association for Computing Machinery, 2020, p. 105–114. [Online].
  Available: \url{https://doi.org/10.1145/3340531.3412003}
\BIBentrySTDinterwordspacing

\bibitem[Mohammad(2012{\natexlab{b}})]{mohammad:2012:STARSEM-SEMEVAL}
\BIBentryALTinterwordspacing
S.~Mohammad, ``\#emotional tweets,'' in \emph{{*SEM 2012}: The First Joint
  Conference on Lexical and Computational Semantics -- Volume 1: Proceedings of
  the main conference and the shared task, and Volume 2: Proceedings of the
  Sixth International Workshop on Semantic Evaluation {(SemEval 2012)}}.\hskip
  1em plus 0.5em minus 0.4em\relax Montr\'eal, Canada: Association for
  Computational Linguistics, 7-8 June 2012, pp. 246--255. [Online]. Available:
  \url{http://www.aclweb.org/anthology/S12-1033}
\BIBentrySTDinterwordspacing

\bibitem[Mohammad and Kiritchenko(2015)]{mohammad2015using}
S.~M. Mohammad and S.~Kiritchenko, ``Using hashtags to capture fine emotion
  categories from tweets,'' \emph{Computational Intelligence}, vol.~31, no.~2,
  pp. 301--326, 2015.

\bibitem[Mohammad(2018{\natexlab{a}})]{mohammad-2018-obtaining}
\BIBentryALTinterwordspacing
S.~Mohammad, ``Obtaining reliable human ratings of valence, arousal, and
  dominance for 20,000 {E}nglish words,'' in \emph{Proceedings of ACL
  2018}.\hskip 1em plus 0.5em minus 0.4em\relax Melbourne, Australia:
  Association for Computational Linguistics, Jul. 2018, pp. 174--184. [Online].
  Available: \url{https://www.aclweb.org/anthology/P18-1017}
\BIBentrySTDinterwordspacing

\bibitem[Mohammad(2018{\natexlab{b}})]{LREC18-AIL}
S.~M. Mohammad, ``Word affect intensities,'' in \emph{Proceedings of LREC
  2018}, Miyazaki, Japan, 2018.

\bibitem[Katz et~al.(2007)Katz, Singleton, and
  Wicentowski]{katz-etal-2007-swat}
\BIBentryALTinterwordspacing
P.~Katz, M.~Singleton, and R.~Wicentowski, ``{SWAT}-{MP}:the {S}em{E}val-2007
  systems for task 5 and task 14,'' in \emph{Proceedings of SemEval
  2007}.\hskip 1em plus 0.5em minus 0.4em\relax Prague, Czech Republic:
  Association for Computational Linguistics, Jun. 2007, pp. 308--313. [Online].
  Available: \url{https://www.aclweb.org/anthology/S07-1067}
\BIBentrySTDinterwordspacing

\bibitem[Bradley and Lang(1994)]{SAM1994}
\BIBentryALTinterwordspacing
M.~M. Bradley and P.~J. Lang, ``Measuring emotion: The self-assessment manikin
  and the semantic differential,'' \emph{Journal of Behavior Therapy and
  Experimental Psychiatry}, vol.~25, no.~1, pp. 49 -- 59, 1994. [Online].
  Available:
  \url{http://www.sciencedirect.com/science/article/pii/0005791694900639}
\BIBentrySTDinterwordspacing

\bibitem[{Otter} et~al.(2020){Otter}, {Medina}, and {Kalita}]{Otter2020_survey}
D.~W. {Otter}, J.~R. {Medina}, and J.~K. {Kalita}, ``A survey of the usages of
  deep learning for natural language processing,'' \emph{IEEE Transactions on
  Neural Networks and Learning Systems}, pp. 1--21, 2020.

\bibitem[Bena and Kalita(2019)]{bena-kalita-2019-introducing}
\BIBentryALTinterwordspacing
B.~Bena and J.~Kalita, ``Introducing aspects of creativity in automatic poetry
  generation,'' in \emph{Proceedings of the 16th International Conference on
  Natural Language Processing}.\hskip 1em plus 0.5em minus 0.4em\relax
  International Institute of Information Technology, Hyderabad, India: NLP
  Association of India, Dec. 2019, pp. 26--35. [Online]. Available:
  \url{https://aclanthology.org/2019.icon-1.4}
\BIBentrySTDinterwordspacing

\bibitem[Ghosh et~al.(2017)Ghosh, Chollet, Laksana, Morency, and
  Scherer]{ghosh-etal-2017-affect}
\BIBentryALTinterwordspacing
S.~Ghosh, M.~Chollet, E.~Laksana, L.-P. Morency, and S.~Scherer, ``Affect-{LM}:
  A neural language model for customizable affective text generation,'' in
  \emph{Proceedings of ACL 2017}.\hskip 1em plus 0.5em minus 0.4em\relax
  Vancouver, Canada: Association for Computational Linguistics, Jul. 2017, pp.
  634--642. [Online]. Available:
  \url{https://www.aclweb.org/anthology/P17-1059}
\BIBentrySTDinterwordspacing

\bibitem[Zhou et~al.(2018)Zhou, Huang, Zhang, Zhu, and
  Liu]{Zhou_2018_AAAI_EmotionalChattingMachine}
\BIBentryALTinterwordspacing
H.~Zhou, M.~Huang, T.~Zhang, X.~Zhu, and B.~Liu, ``Emotional chatting machine:
  Emotional conversation generation with internal and external memory,'' 2018.
  [Online]. Available:
  \url{https://aaai.org/ocs/index.php/AAAI/AAAI18/paper/view/16455}
\BIBentrySTDinterwordspacing

\bibitem[Ma et~al.(2020)Ma, Nguyen, Xing, and Cambria]{MA202050}
\BIBentryALTinterwordspacing
Y.~Ma, K.~L. Nguyen, F.~Z. Xing, and E.~Cambria, ``A survey on empathetic
  dialogue systems,'' \emph{Information Fusion}, vol.~64, pp. 50 -- 70, 2020.
  [Online]. Available:
  \url{http://www.sciencedirect.com/science/article/pii/S1566253520303092}
\BIBentrySTDinterwordspacing

\bibitem[Peng et~al.(2020)Peng, Zhou, Liu, and Ai]{PENG2020105319}
\BIBentryALTinterwordspacing
D.~Peng, M.~Zhou, C.~Liu, and J.~Ai, ``Human–machine dialogue modelling with
  the fusion of word- and sentence-level emotions,'' \emph{Knowledge-Based
  Systems}, vol. 192, p. 105319, 2020. [Online]. Available:
  \url{http://www.sciencedirect.com/science/article/pii/S0950705119305970}
\BIBentrySTDinterwordspacing

\bibitem[Song et~al.(2019)Song, Zheng, Liu, Xu, and
  Huang]{song-etal-2019-generating}
\BIBentryALTinterwordspacing
Z.~Song, X.~Zheng, L.~Liu, M.~Xu, and X.~Huang, ``Generating responses with a
  specific emotion in dialog,'' in \emph{Proceedings of ACL 2019}.\hskip 1em
  plus 0.5em minus 0.4em\relax Florence, Italy: Association for Computational
  Linguistics, Jul. 2019, pp. 3685--3695. [Online]. Available:
  \url{https://www.aclweb.org/anthology/P19-1359}
\BIBentrySTDinterwordspacing

\bibitem[Klein et~al.(1973)Klein, Aeschlimann, Balsiger, Converse, Court,
  Foster, Lao, Oakley, and Smith]{Klein1973AutomaticNW}
S.~Klein, J.~F. Aeschlimann, D.~F. Balsiger, S.~L. Converse, C.~Court,
  M.~Foster, R.~Lao, J.~Oakley, and J.~Smith, ``Automatic novel writing: A
  status report,'' 1973.

\bibitem[Meehan(1976)]{meehan1976metanovel}
J.~R. Meehan, ``The metanovel: writing stories by computer.'' YALE UNIV NEW
  HAVEN CONN DEPT OF COMPUTER SCIENCE, Tech. Rep., 1976.

\bibitem[Meehan(1977)]{Meehan_1977_IJCAI}
------, ``Tale-spin, an interactive program that writes stories,'' in
  \emph{Proceedings of IJCAI}.\hskip 1em plus 0.5em minus 0.4em\relax San
  Francisco, CA, USA: Morgan Kaufmann Publishers Inc., 1977, p. 91–98.

\bibitem[Lebowitz(1984)]{LEBOWITZ1984171}
\BIBentryALTinterwordspacing
M.~Lebowitz, ``Creating characters in a story-telling universe,''
  \emph{Poetics}, vol.~13, no.~3, pp. 171 -- 194, 1984. [Online]. Available:
  \url{http://www.sciencedirect.com/science/article/pii/0304422X84900019}
\BIBentrySTDinterwordspacing

\bibitem[Lebowitz(1985)]{LEBOWITZ1985483}
\BIBentryALTinterwordspacing
------, ``Story-telling as planning and learning,'' \emph{Poetics}, vol.~14,
  no.~6, pp. 483 -- 502, 1985. [Online]. Available:
  \url{http://www.sciencedirect.com/science/article/pii/0304422X85900154}
\BIBentrySTDinterwordspacing

\bibitem[Gerv{á}s(2009)]{Gervas_2009_storytelling_and_creativity}
P.~Gerv{á}s, ``Computational approaches to storytelling and creativity,''
  \emph{AI Magazine}, vol.~30, pp. 49--62, 09 2009.

\bibitem[McIntyre and Lapata(2009)]{mcintyre-lapata-2009-learning}
\BIBentryALTinterwordspacing
N.~McIntyre and M.~Lapata, ``Learning to tell tales: A data-driven approach to
  story generation,'' in \emph{Proceedings of ACL-AFNLP 2009}.\hskip 1em plus
  0.5em minus 0.4em\relax Suntec, Singapore: Association for Computational
  Linguistics, Aug. 2009, pp. 217--225. [Online]. Available:
  \url{https://www.aclweb.org/anthology/P09-1025}
\BIBentrySTDinterwordspacing

\bibitem[Hermann et~al.(2015)Hermann, Ko{\v{c}}isk{\'{y}}, Grefenstette,
  Espeholt, Kay, Suleyman, and Blunsom]{Hermann:2015:TMR:2969239.2969428}
\BIBentryALTinterwordspacing
K.~M. Hermann, T.~Ko{\v{c}}isk{\'{y}}, E.~Grefenstette, L.~Espeholt, W.~Kay,
  M.~Suleyman, and P.~Blunsom, ``Teaching machines to read and comprehend,'' in
  \emph{Proceedings of NIPS}.\hskip 1em plus 0.5em minus 0.4em\relax Cambridge,
  MA, USA: MIT Press, 2015, pp. 1693--1701. [Online]. Available:
  \url{http://dl.acm.org/citation.cfm?id=2969239.2969428}
\BIBentrySTDinterwordspacing

\bibitem[Roemmele(2016)]{roemmele_writing_2016}
\BIBentryALTinterwordspacing
M.~Roemmele, ``Writing {Stories} with {Help} from {Recurrent} {Neural}
  {Networks},'' in \emph{Proceedings of AAAI}.\hskip 1em plus 0.5em minus
  0.4em\relax Phoenix, AZ: AAAI Press, Feb. 2016, pp. 4311 -- 4312. [Online].
  Available:
  \url{http://www.aaai.org/ocs/index.php/AAAI/AAAI16/paper/view/11966}
\BIBentrySTDinterwordspacing

\bibitem[Peng et~al.(2018)Peng, Ghazvininejad, May, and
  Knight]{peng2018towards}
\BIBentryALTinterwordspacing
N.~Peng, M.~Ghazvininejad, J.~May, and K.~Knight, ``Towards controllable story
  generation,'' in \emph{Proceedings of the First Workshop on
  Storytelling}.\hskip 1em plus 0.5em minus 0.4em\relax New Orleans, Louisiana:
  Association for Computational Linguistics, Jun. 2018, pp. 43--49. [Online].
  Available: \url{https://www.aclweb.org/anthology/W18-1505}
\BIBentrySTDinterwordspacing

\bibitem[Yao et~al.(2019)Yao, Peng, Ralph, Knight, Zhao, and Yan]{yao2019plan}
\BIBentryALTinterwordspacing
L.~Yao, N.~Peng, W.~Ralph, K.~Knight, D.~Zhao, and R.~Yan, ``Plan-and-{W}rite:
  Towards better automatic storytelling,'' in \emph{Proceedings of AAAI}.\hskip
  1em plus 0.5em minus 0.4em\relax Honolulu, Hawaii: {AAAI} Press,
  January--February 2019, pp. 7378--7385. [Online]. Available:
  \url{https://doi.org/10.1609/aaai.v33i01.33017378}
\BIBentrySTDinterwordspacing

\bibitem[Goldfarb-Tarrant et~al.(2019)Goldfarb-Tarrant, Feng, and
  Peng]{goldfarb2019plan}
\BIBentryALTinterwordspacing
S.~Goldfarb-Tarrant, H.~Feng, and N.~Peng, ``{P}lan, {W}rite, and {R}evise: an
  interactive system for open-domain story generation,'' in \emph{Proceedings
  of NAACL-HLT 2019 (Demonstrations)}.\hskip 1em plus 0.5em minus 0.4em\relax
  Minneapolis, Minnesota: Association for Computational Linguistics, Jun. 2019,
  pp. 89--97. [Online]. Available:
  \url{https://www.aclweb.org/anthology/N19-4016}
\BIBentrySTDinterwordspacing

\bibitem[Hou et~al.(2019)Hou, Zhou, Zhou, Sun, and
  Xuanyuan]{Hou_2019_survey_dl_storygeneration}
C.~Hou, C.~Zhou, K.~Zhou, J.~Sun, and S.~Xuanyuan, ``A survey of deep learning
  applied to story generation,'' in \emph{Smart Computing and Communication},
  M.~Qiu, Ed.\hskip 1em plus 0.5em minus 0.4em\relax Cham: Springer
  International Publishing, 2019, pp. 1--10.

\bibitem[Fan et~al.(2018)Fan, Lewis, and Dauphin]{WritingPromptsDataset}
\BIBentryALTinterwordspacing
A.~Fan, M.~Lewis, and Y.~Dauphin, ``Hierarchical neural story generation,'' in
  \emph{Proceedings of ACL 2018}.\hskip 1em plus 0.5em minus 0.4em\relax
  Melbourne, Australia: Association for Computational Linguistics, July 2018,
  pp. 889--898. [Online]. Available:
  \url{http://www.aclweb.org/anthology/P18-1082}
\BIBentrySTDinterwordspacing

\bibitem[Xu et~al.(2018)Xu, Ren, Zhang, Zeng, Cai, and
  Sun]{xu-etal-2018-skeleton}
\BIBentryALTinterwordspacing
J.~Xu, X.~Ren, Y.~Zhang, Q.~Zeng, X.~Cai, and X.~Sun, ``A skeleton-based model
  for promoting coherence among sentences in narrative story generation,'' in
  \emph{Proceedings of EMNLP 2018}.\hskip 1em plus 0.5em minus 0.4em\relax
  Brussels, Belgium: Association for Computational Linguistics, Oct.-Nov. 2018,
  pp. 4306--4315. [Online]. Available:
  \url{https://www.aclweb.org/anthology/D18-1462}
\BIBentrySTDinterwordspacing

\bibitem[Wang et~al.(2018)Wang, Chen, Wang, and Wang]{wang2018no}
X.~Wang, W.~Chen, Y.-F. Wang, and W.~Y. Wang, ``No metrics are perfect:
  Adversarial reward learning for visual storytelling,'' in \emph{Proceedings
  of ACL 2018}, 2018, pp. 899--909.

\bibitem[Zhao et~al.(2018)Zhao, Liu, Liu, Yang, and Yu]{Zhao2018plotstoendings}
\BIBentryALTinterwordspacing
Y.~Zhao, L.~Liu, C.~Liu, R.~Yang, and D.~Yu, ``From plots to endings: {A}
  reinforced pointer generator for story ending generation,'' in
  \emph{Proceedings of Natural Language Processing and Chinese Computing}, vol.
  abs/1901.03459, 2018. [Online]. Available:
  \url{http://arxiv.org/abs/1901.03459}
\BIBentrySTDinterwordspacing

\bibitem[Wang and Wan(2019)]{wang_tcvae}
\BIBentryALTinterwordspacing
T.~Wang and X.~Wan, ``{T-CVAE}: Transformer-based conditioned variational
  autoencoder for story completion,'' in \emph{Proceedings of IJCAI}.\hskip 1em
  plus 0.5em minus 0.4em\relax International Joint Conferences on Artificial
  Intelligence Organization, July 2019, pp. 5233--5239. [Online]. Available:
  \url{https://doi.org/10.24963/ijcai.2019/727}
\BIBentrySTDinterwordspacing

\bibitem[Clark et~al.(2018{\natexlab{a}})Clark, Ross, Tan, Ji, and
  Smith]{Clark_2018_creative_Writing_machine_in_the_loop}
\BIBentryALTinterwordspacing
E.~Clark, A.~S. Ross, C.~Tan, Y.~Ji, and N.~A. Smith, ``Creative writing with a
  machine in the loop: Case studies on slogans and stories,'' in
  \emph{Proceedings of IUI}.\hskip 1em plus 0.5em minus 0.4em\relax New York,
  NY, USA: Association for Computing Machinery, 2018, p. 329–340. [Online].
  Available: \url{https://doi.org/10.1145/3172944.3172983}
\BIBentrySTDinterwordspacing

\bibitem[Li et~al.(2013)Li, Lee-Urban, Johnston, and Riedl]{Li_2013_AAAI}
\BIBentryALTinterwordspacing
B.~Li, S.~Lee-Urban, G.~Johnston, and M.~Riedl, ``Story generation with
  crowdsourced plot graphs,'' 2013. [Online]. Available:
  \url{https://www.aaai.org/ocs/index.php/AAAI/AAAI13/paper/view/6399}
\BIBentrySTDinterwordspacing

\bibitem[Fan et~al.(2019)Fan, Lewis, and Dauphin]{fan-etal-2019-strategies}
\BIBentryALTinterwordspacing
A.~Fan, M.~Lewis, and Y.~Dauphin, ``Strategies for structuring story
  generation,'' in \emph{Proceedings of ACL 2019}.\hskip 1em plus 0.5em minus
  0.4em\relax Florence, Italy: Association for Computational Linguistics, Jul.
  2019, pp. 2650--2660. [Online]. Available:
  \url{https://www.aclweb.org/anthology/P19-1254}
\BIBentrySTDinterwordspacing

\bibitem[Goldfarb-Tarrant et~al.(2020)Goldfarb-Tarrant, Chakrabarty,
  Weischedel, and Peng]{goldfarb2020content}
S.~Goldfarb-Tarrant, T.~Chakrabarty, R.~Weischedel, and N.~Peng, ``Content
  planning for neural story generation with aristotelian rescoring,'' in
  \emph{Proceedings of EMNLP 2020}, 2020.

\bibitem[Holtzman et~al.(2018)Holtzman, Buys, Forbes, Bosselut, Golub, and
  Choi]{holtzman-etal-2018-learning}
\BIBentryALTinterwordspacing
A.~Holtzman, J.~Buys, M.~Forbes, A.~Bosselut, D.~Golub, and Y.~Choi, ``Learning
  to write with cooperative discriminators,'' in \emph{Proceedings of ACL
  2018}.\hskip 1em plus 0.5em minus 0.4em\relax Melbourne, Australia:
  Association for Computational Linguistics, Jul. 2018, pp. 1638--1649.
  [Online]. Available: \url{https://www.aclweb.org/anthology/P18-1152}
\BIBentrySTDinterwordspacing

\bibitem[Vaswani et~al.(2017)Vaswani, Shazeer, Parmar, Uszkoreit, Jones, Gomez,
  Kaiser, and Polosukhin]{Vaswani_2017_transformer}
\BIBentryALTinterwordspacing
A.~Vaswani, N.~Shazeer, N.~Parmar, J.~Uszkoreit, L.~Jones, A.~N. Gomez, L.~u.
  Kaiser, and I.~Polosukhin, ``Attention is all you need,'' in
  \emph{Proceedings of NIPS}, I.~Guyon, U.~V. Luxburg, S.~Bengio, H.~Wallach,
  R.~Fergus, S.~Vishwanathan, and R.~Garnett, Eds.\hskip 1em plus 0.5em minus
  0.4em\relax Curran Associates, Inc., 2017, pp. 5998--6008. [Online].
  Available:
  \url{http://papers.nips.cc/paper/7181-attention-is-all-you-need.pdf}
\BIBentrySTDinterwordspacing

\bibitem[Sutskever et~al.(2014)Sutskever, Vinyals, and
  Le]{Sutskever:2014:SSL:2969033.2969173}
\BIBentryALTinterwordspacing
I.~Sutskever, O.~Vinyals, and Q.~V. Le, ``Sequence to sequence learning with
  neural networks,'' in \emph{Proceedings of NIPS}.\hskip 1em plus 0.5em minus
  0.4em\relax Cambridge, MA, USA: MIT Press, 2014, pp. 3104--3112. [Online].
  Available: \url{http://dl.acm.org/citation.cfm?id=2969033.2969173}
\BIBentrySTDinterwordspacing

\bibitem[Vinyals and Le(2015)]{Vinyals2015conversation}
\BIBentryALTinterwordspacing
O.~Vinyals and Q.~V. Le, ``A neural conversational model,'' in
  \emph{Proceedings of the 31st ICML Deep Learning Workshop}, 2015. [Online].
  Available: \url{http://arxiv.org/pdf/1506.05869v3.pdf}
\BIBentrySTDinterwordspacing

\bibitem[Gehring et~al.(2017)Gehring, Auli, Grangier, Yarats, and
  Dauphin]{pmlr-v70-gehring17a}
\BIBentryALTinterwordspacing
J.~Gehring, M.~Auli, D.~Grangier, D.~Yarats, and Y.~N. Dauphin, ``Convolutional
  sequence to sequence learning,'' ser. Proceedings of Machine Learning
  Research, D.~Precup and Y.~W. Teh, Eds., vol.~70.\hskip 1em plus 0.5em minus
  0.4em\relax International Convention Centre, Sydney, Australia: PMLR, 06--11
  Aug 2017, pp. 1243--1252. [Online]. Available:
  \url{http://proceedings.mlr.press/v70/gehring17a.html}
\BIBentrySTDinterwordspacing

\bibitem[ten(Last updated 2020-10-14 UTC.)]{tensorflow_transformer_tutorial}
``Transformer model for language understanding,''
  \url{https://www.tensorflow.org/tutorials/text/transformer}, Last updated
  2020-10-14 UTC., accessed: October 19, 2020.

\bibitem[Radford et~al.(2019)Radford, Wu, Child, Luan, Amodei, and
  Sutskever]{radford2019language}
\BIBentryALTinterwordspacing
A.~Radford, J.~Wu, R.~Child, D.~Luan, D.~Amodei, and I.~Sutskever, ``Language
  models are unsupervised multitask learners,'' 2019. [Online]. Available:
  \url{https://d4mucfpksywv.cloudfront.net/better-language-models/language-model.pdf}
\BIBentrySTDinterwordspacing

\bibitem[Yang et~al.(2019)Yang, Dai, Yang, Carbonell, Salakhutdinov, and
  Le]{Yang_NIPS2019_8812}
\BIBentryALTinterwordspacing
Z.~Yang, Z.~Dai, Y.~Yang, J.~Carbonell, R.~R. Salakhutdinov, and Q.~V. Le,
  ``{XL}{N}et: Generalized autoregressive pretraining for language
  understanding,'' in \emph{Proceedings of NeurIPS}, H.~Wallach, H.~Larochelle,
  A.~Beygelzimer, F.~d\textquotesingle Alch\'{e}-Buc, E.~Fox, and R.~Garnett,
  Eds.\hskip 1em plus 0.5em minus 0.4em\relax Curran Associates, Inc., 2019,
  pp. 5753--5763. [Online]. Available:
  \url{http://papers.nips.cc/paper/8812-xlnet-generalized-autoregressive-pretraining-for-language-understanding.pdf}
\BIBentrySTDinterwordspacing

\bibitem[Lewis et~al.(2020)Lewis, Liu, Goyal, Ghazvininejad, Mohamed, Levy,
  Stoyanov, and Zettlemoyer]{lewis-etal-2020-bart}
\BIBentryALTinterwordspacing
M.~Lewis, Y.~Liu, N.~Goyal, M.~Ghazvininejad, A.~Mohamed, O.~Levy, V.~Stoyanov,
  and L.~Zettlemoyer, ``{BART}: Denoising sequence-to-sequence pre-training for
  natural language generation, translation, and comprehension,'' in
  \emph{Proceedings of ACL 2020}.\hskip 1em plus 0.5em minus 0.4em\relax
  Online: Association for Computational Linguistics, Jul. 2020, pp. 7871--7880.
  [Online]. Available: \url{https://www.aclweb.org/anthology/2020.acl-main.703}
\BIBentrySTDinterwordspacing

\bibitem[Rothe et~al.(2020)Rothe, Narayan, and Severyn]{Rothe2020}
S.~Rothe, S.~Narayan, and A.~Severyn, ``Leveraging pre-trained checkpoints for
  sequence generation tasks,'' \emph{TACL}, vol.~8, pp. 264--280, 2020.

\bibitem[Raffel et~al.(2020)Raffel, Shazeer, Roberts, Lee, Narang, Matena,
  Zhou, Li, and Liu]{2020t5}
\BIBentryALTinterwordspacing
C.~Raffel, N.~Shazeer, A.~Roberts, K.~Lee, S.~Narang, M.~Matena, Y.~Zhou,
  W.~Li, and P.~J. Liu, ``Exploring the limits of transfer learning with a
  unified text-to-text transformer,'' \emph{Journal of Machine Learning
  Research}, vol.~21, no. 140, pp. 1--67, 2020. [Online]. Available:
  \url{http://jmlr.org/papers/v21/20-074.html}
\BIBentrySTDinterwordspacing

\bibitem[Qi et~al.(2020)Qi, Yan, Gong, Liu, Duan, Chen, Zhang, and
  Zhou]{qi2020prophetnet}
W.~Qi, Y.~Yan, Y.~Gong, D.~Liu, N.~Duan, J.~Chen, R.~Zhang, and M.~Zhou,
  ``Prophetnet: Predicting future n-gram for sequence-to-sequence
  pre-training,'' in \emph{Proceedings of EMNLP 2020}.\hskip 1em plus 0.5em
  minus 0.4em\relax Punta Cana, Dominican Republic: Association for
  Computational Linguistics, November 2020.

\bibitem[Roller et~al.(2020)Roller, Dinan, Goyal, Ju, Williamson, Liu, Xu, Ott,
  Shuster, Smith, Boureau, and Weston]{roller2020recipes}
S.~Roller, E.~Dinan, N.~Goyal, D.~Ju, M.~Williamson, Y.~Liu, J.~Xu, M.~Ott,
  K.~Shuster, E.~M. Smith, Y.-L. Boureau, and J.~Weston, ``Recipes for building
  an open-domain chatbot,'' 2020.

\bibitem[Zhang et~al.(2020{\natexlab{a}})Zhang, Zhao, Saleh, and
  Liu]{zhang2020pegasus}
J.~Zhang, Y.~Zhao, M.~Saleh, and P.~J. Liu, ``Pegasus: Pre-training with
  extracted gap-sentences for abstractive summarization,'' 2020.

\bibitem[Keskar et~al.(2019)Keskar, McCann, Varshney, Xiong, and
  Socher]{keskar2019ctrl}
N.~S. Keskar, B.~McCann, L.~R. Varshney, C.~Xiong, and R.~Socher, ``Ctrl: A
  conditional transformer language model for controllable generation,''
  \emph{arXiv preprint arXiv:1909.05858}, 2019.

\bibitem[Shoeybi et~al.(2020)Shoeybi, Patwary, Puri, LeGresley, Casper, and
  Catanzaro]{shoeybi2020megatronlm}
M.~Shoeybi, M.~Patwary, R.~Puri, P.~LeGresley, J.~Casper, and B.~Catanzaro,
  ``Megatron-lm: Training multi-billion parameter language models using model
  parallelism,'' 2020.

\bibitem[Guan et~al.(2020)Guan, Huang, Zhao, Zhu, and Huang]{Guan_2020_TACL}
\BIBentryALTinterwordspacing
J.~Guan, F.~Huang, Z.~Zhao, X.~Zhu, and M.~Huang, ``A knowledge-enhanced
  pretraining model for commonsense story generation,'' \emph{TACL}, vol.~8,
  pp. 93--108, 2020. [Online]. Available:
  \url{https://doi.org/10.1162/tacl_a_00302}
\BIBentrySTDinterwordspacing

\bibitem[Rashkin et~al.(2020)Rashkin, Celikyilmaz, Choi, and
  Gao]{rashkin2020plotmachines}
H.~Rashkin, A.~Celikyilmaz, Y.~Choi, and J.~Gao, ``Plotmachines:
  Outline-conditioned generation with dynamic plot state tracking,'' 2020.

\bibitem[Nguyen et~al.(2017)Nguyen, Clune, Bengio, Dosovitskiy, and
  Yosinski]{nguyen2017plug}
A.~Nguyen, J.~Clune, Y.~Bengio, A.~Dosovitskiy, and J.~Yosinski, ``Plug \& play
  generative networks: Conditional iterative generation of images in latent
  space,'' in \emph{Proceedings of the IEEE Conference on Computer Vision and
  Pattern Recognition}.\hskip 1em plus 0.5em minus 0.4em\relax IEEE, 2017.

\bibitem[Qin et~al.(2020)Qin, Shwartz, West, Bhagavatula, Hwang, Le~Bras,
  Bosselut, and Choi]{qin-etal-2020-back}
\BIBentryALTinterwordspacing
L.~Qin, V.~Shwartz, P.~West, C.~Bhagavatula, J.~D. Hwang, R.~Le~Bras,
  A.~Bosselut, and Y.~Choi, ``Back to the future: Unsupervised backprop-based
  decoding for counterfactual and abductive commonsense reasoning,'' in
  \emph{Proceedings of EMNLP 2020}.\hskip 1em plus 0.5em minus 0.4em\relax
  Online: Association for Computational Linguistics, Nov. 2020, pp. 794--805.
  [Online]. Available:
  \url{https://www.aclweb.org/anthology/2020.emnlp-main.58}
\BIBentrySTDinterwordspacing

\bibitem[Zhang et~al.(2020{\natexlab{b}})Zhang, Sax, Zamir, Guibas, and
  Malik]{Zhang2020_Side-Tuning}
J.~O. Zhang, A.~Sax, A.~Zamir, L.~Guibas, and J.~Malik, ``Side-tuning: A
  baseline for network adaptation via additive side networks,'' in
  \emph{Proceedings of ECCV}, A.~Vedaldi, H.~Bischof, T.~Brox, and J.-M. Frahm,
  Eds.\hskip 1em plus 0.5em minus 0.4em\relax Cham: Springer International
  Publishing, 2020, pp. 698--714.

\bibitem[Zeldes et~al.(2020)Zeldes, Padnos, Sharir, and
  Peleg]{zeldes2020technical}
Y.~Zeldes, D.~Padnos, O.~Sharir, and B.~Peleg, ``Technical report: Auxiliary
  tuning and its application to conditional text generation,'' 2020.

\bibitem[Krause et~al.(2021)Krause, Gotmare, McCann, Keskar, Joty, richard
  socher, and Rajani]{krause2021gedi}
\BIBentryALTinterwordspacing
B.~Krause, A.~D. Gotmare, B.~McCann, N.~S. Keskar, S.~Joty, richard socher, and
  N.~Rajani, ``Gedi: Generative discriminator guided sequence generation,''
  2021. [Online]. Available: \url{https://openreview.net/forum?id=TJSOfuZEd1B}
\BIBentrySTDinterwordspacing

\bibitem[Ippolito et~al.(2019)Ippolito, Grangier, Callison-Burch, and
  Eck]{ippolito-etal-2019-unsupervised}
\BIBentryALTinterwordspacing
D.~Ippolito, D.~Grangier, C.~Callison-Burch, and D.~Eck, ``Unsupervised
  hierarchical story infilling,'' in \emph{Proceedings of the First Workshop on
  Narrative Understanding}.\hskip 1em plus 0.5em minus 0.4em\relax Minneapolis,
  Minnesota: Association for Computational Linguistics, Jun. 2019, pp. 37--43.
  [Online]. Available: \url{https://www.aclweb.org/anthology/W19-2405}
\BIBentrySTDinterwordspacing

\bibitem[August et~al.(2020)August, Sap, Clark, Reinecke, and
  Smith]{august-etal-2020-exploring}
\BIBentryALTinterwordspacing
T.~August, M.~Sap, E.~Clark, K.~Reinecke, and N.~A. Smith, ``Exploring the
  effect of author and reader identity in online story writing: the
  {STORIESINTHEWILD} corpus.'' in \emph{Proceedings of the First Joint Workshop
  on Narrative Understanding, Storylines, and Events}.\hskip 1em plus 0.5em
  minus 0.4em\relax Online: Association for Computational Linguistics, Jul.
  2020, pp. 46--54. [Online]. Available:
  \url{https://www.aclweb.org/anthology/2020.nuse-1.6}
\BIBentrySTDinterwordspacing

\bibitem[Liu et~al.(2016)Liu, Lowe, Serban, Noseworthy, Charlin, and
  Pineau]{liu-EtAl:2016:EMNLP20163}
\BIBentryALTinterwordspacing
C.-W. Liu, R.~Lowe, I.~Serban, M.~Noseworthy, L.~Charlin, and J.~Pineau, ``How
  not to evaluate your dialogue system: An empirical study of unsupervised
  evaluation metrics for dialogue response generation,'' in \emph{Proceedings
  of EMNLP 2016}.\hskip 1em plus 0.5em minus 0.4em\relax Austin, Texas:
  Association for Computational Linguistics, November 2016, pp. 2122--2132.
  [Online]. Available: \url{https://aclweb.org/anthology/D16-1230}
\BIBentrySTDinterwordspacing

\bibitem[Novikova et~al.(2017)Novikova, Du{\v{s}}ek, Cercas~Curry, and
  Rieser]{novikova-etal-2017-need}
\BIBentryALTinterwordspacing
J.~Novikova, O.~Du{\v{s}}ek, A.~Cercas~Curry, and V.~Rieser, ``Why we need new
  evaluation metrics for {NLG},'' in \emph{Proceedings of EMNLP 2017}.\hskip
  1em plus 0.5em minus 0.4em\relax Copenhagen, Denmark: Association for
  Computational Linguistics, Sep. 2017, pp. 2241--2252. [Online]. Available:
  \url{https://www.aclweb.org/anthology/D17-1238}
\BIBentrySTDinterwordspacing

\bibitem[Chaganty et~al.(2018)Chaganty, Mussmann, and
  Liang]{chaganty-etal-2018-price}
\BIBentryALTinterwordspacing
A.~Chaganty, S.~Mussmann, and P.~Liang, ``The price of debiasing automatic
  metrics in natural language evalaution,'' in \emph{Proceedings of ACL
  2018}.\hskip 1em plus 0.5em minus 0.4em\relax Melbourne, Australia:
  Association for Computational Linguistics, Jul. 2018, pp. 643--653. [Online].
  Available: \url{https://www.aclweb.org/anthology/P18-1060}
\BIBentrySTDinterwordspacing

\bibitem[Gatt and Krahmer(2018)]{Gatt_and_Krahmer_suvery_NLG}
A.~Gatt and E.~Krahmer, ``\BIBforeignlanguage{English}{Survey of the state of
  the art in natural language generation: Core tasks, applications and
  evaluation},'' \emph{\BIBforeignlanguage{English}{Journal of Artificial
  Intelligence Research}}, vol.~61, no.~1, pp. 65--170, 2018.

\bibitem[Hashimoto et~al.(2019)Hashimoto, Zhang, and
  Liang]{hashimoto-etal-2019-unifying}
\BIBentryALTinterwordspacing
T.~Hashimoto, H.~Zhang, and P.~Liang, ``Unifying human and statistical
  evaluation for natural language generation,'' in \emph{Proceedings of
  NAACL-HLT 2019}.\hskip 1em plus 0.5em minus 0.4em\relax Minneapolis,
  Minnesota: Association for Computational Linguistics, Jun. 2019, pp.
  1689--1701. [Online]. Available:
  \url{https://www.aclweb.org/anthology/N19-1169}
\BIBentrySTDinterwordspacing

\bibitem[Cavazza et~al.(2009)Cavazza, Pizzi, Charles, Vogt, and
  Andr\'{e}]{Cavazza_2009_emotional_input}
M.~Cavazza, D.~Pizzi, F.~Charles, T.~Vogt, and E.~Andr\'{e}, ``Emotional input
  for character-based interactive storytelling,'' in \emph{Proceedings of
  AAMAS}.\hskip 1em plus 0.5em minus 0.4em\relax Richland, SC: International
  Foundation for Autonomous Agents and Multiagent Systems, 2009, p. 313–320.

\bibitem[Shuster et~al.(2020)Shuster, Humeau, Bordes, and
  Weston]{shuster-etal-2020-image}
\BIBentryALTinterwordspacing
K.~Shuster, S.~Humeau, A.~Bordes, and J.~Weston, ``Image-chat: Engaging
  grounded conversations,'' in \emph{Proceedings of ACL 2020}.\hskip 1em plus
  0.5em minus 0.4em\relax Online: Association for Computational Linguistics,
  Jul. 2020, pp. 2414--2429. [Online]. Available:
  \url{https://www.aclweb.org/anthology/2020.acl-main.219}
\BIBentrySTDinterwordspacing

\bibitem[Donahue et~al.(2020)Donahue, Lee, and
  Liang]{donahue-etal-2020-enabling}
\BIBentryALTinterwordspacing
C.~Donahue, M.~Lee, and P.~Liang, ``Enabling language models to fill in the
  blanks,'' in \emph{Proceedings of ACL 2020}.\hskip 1em plus 0.5em minus
  0.4em\relax Online: Association for Computational Linguistics, Jul. 2020, pp.
  2492--2501. [Online]. Available:
  \url{https://www.aclweb.org/anthology/2020.acl-main.225}
\BIBentrySTDinterwordspacing

\bibitem[Huang et~al.(2020)Huang, Zhang, Elachqar, and
  Cheng]{huang-etal-2020-inset}
\BIBentryALTinterwordspacing
Y.~Huang, Y.~Zhang, O.~Elachqar, and Y.~Cheng, ``{INSET}: Sentence infilling
  with {IN}ter-{SE}ntential transformer,'' in \emph{Proceedings of ACL
  2020}.\hskip 1em plus 0.5em minus 0.4em\relax Online: Association for
  Computational Linguistics, Jul. 2020, pp. 2502--2515. [Online]. Available:
  \url{https://www.aclweb.org/anthology/2020.acl-main.226}
\BIBentrySTDinterwordspacing

\bibitem[Wang et~al.(2020)Wang, Durrett, and Erk]{wang2020narrative}
S.~Wang, G.~Durrett, and K.~Erk, ``Narrative interpolation for generating and
  understanding stories,'' 2020.

\bibitem[Clarke et~al.(2019)Clarke, Braun, Frith, and
  Moller]{Clarke_2019_doi:10.1080/14780887.2018.1536378}
\BIBentryALTinterwordspacing
V.~Clarke, V.~Braun, H.~Frith, and N.~Moller, ``Editorial introduction to the
  special issue: Using story completion methods in qualitative research,''
  \emph{Qualitative Research in Psychology}, vol.~16, no.~1, pp. 1--20, 2019.
  [Online]. Available: \url{https://doi.org/10.1080/14780887.2018.1536378}
\BIBentrySTDinterwordspacing

\bibitem[Roemmele(2021)]{Roemmele2021InspirationTO}
M.~Roemmele, ``Inspiration through observation: Demonstrating the influence of
  automatically generated text on creative writing,'' in \emph{Proceedings of
  ICCC 2021}.\hskip 1em plus 0.5em minus 0.4em\relax Association for
  Computational Creativity {(ACC)}, September 2021.

\bibitem[Bhagavatula et~al.(2020)Bhagavatula, Bras, Malaviya, Sakaguchi,
  Holtzman, Rashkin, Downey, tau Yih, and Choi]{Bhagavatula2020Abductive}
\BIBentryALTinterwordspacing
C.~Bhagavatula, R.~L. Bras, C.~Malaviya, K.~Sakaguchi, A.~Holtzman, H.~Rashkin,
  D.~Downey, W.~tau Yih, and Y.~Choi, ``Abductive commonsense reasoning,'' in
  \emph{Proceedings of ICLR}, 2020. [Online]. Available:
  \url{https://openreview.net/forum?id=Byg1v1HKDB}
\BIBentrySTDinterwordspacing

\bibitem[Hobbs et~al.(1993)Hobbs, Stickel, Appelt, and Martin]{HOBBS199369}
\BIBentryALTinterwordspacing
J.~R. Hobbs, M.~E. Stickel, D.~E. Appelt, and P.~Martin, ``Interpretation as
  abduction,'' \emph{Artificial Intelligence}, vol.~63, no.~1, pp. 69--142,
  1993. [Online]. Available:
  \url{https://www.sciencedirect.com/science/article/pii/0004370293900154}
\BIBentrySTDinterwordspacing

\bibitem[Norvig(1987)]{Norvig1987}
P.~Norvig, ``Inference in text understanding,'' in \emph{Proceedings of
  AAAI}.\hskip 1em plus 0.5em minus 0.4em\relax AAAI Press, 1987, p. 561–565.

\bibitem[Charniak and Shimony(1990)]{Charniak_and_Shimony_1990}
E.~Charniak and S.~E. Shimony, ``Probabilistic semantics for cost based
  abduction,'' in \emph{Proceedings of AAAI}.\hskip 1em plus 0.5em minus
  0.4em\relax AAAI Press, 1990, p. 106–111.

\bibitem[Peirce et~al.(1937)Peirce, Hartshorne, and Weiss]{Peirce1937-PEICPO-7}
C.~S. Peirce, C.~Hartshorne, and P.~Weiss, \emph{Collected Papers of C. S.
  Peirce. Vol. V. Pragmatism and Pragmaticism}.\hskip 1em plus 0.5em minus
  0.4em\relax University of Chicago Press, 1937, vol.~4, no.~1.

\bibitem[Andersen(1973)]{Andersen_1973}
\BIBentryALTinterwordspacing
H.~Andersen, ``Abductive and deductive change,'' \emph{Language}, vol.~49,
  no.~4, pp. 765--793, 1973. [Online]. Available:
  \url{http://www.jstor.org/stable/412063}
\BIBentrySTDinterwordspacing

\bibitem[Pearl(2002)]{Pearl_2002}
\BIBentryALTinterwordspacing
J.~Pearl, ``Reasoning with cause and effect,'' \emph{AI Magazine}, vol.~23,
  no.~1, p.~95, Mar. 2002. [Online]. Available:
  \url{https://ojs.aaai.org/index.php/aimagazine/article/view/1612}
\BIBentrySTDinterwordspacing

\bibitem[Pearl and Mackenzie(2018)]{Pearl_and_Mackenzie_2018}
J.~Pearl and D.~Mackenzie, \emph{The Book of Why: The New Science of Cause and
  Effect}, 1st~ed.\hskip 1em plus 0.5em minus 0.4em\relax USA: Basic Books,
  Inc., 2018.

\bibitem[Zhu et~al.(2020)Zhu, Song, Dou, NIE, and
  Zhou]{zhu-etal-2020-scriptwriter}
\BIBentryALTinterwordspacing
Y.~Zhu, R.~Song, Z.~Dou, J.-Y. NIE, and J.~Zhou, ``{S}cript{W}riter:
  Narrative-guided script generation,'' in \emph{Proceedings of ACL
  2020}.\hskip 1em plus 0.5em minus 0.4em\relax Online: Association for
  Computational Linguistics, Jul. 2020, pp. 8647--8657. [Online]. Available:
  \url{https://www.aclweb.org/anthology/2020.acl-main.765}
\BIBentrySTDinterwordspacing

\bibitem[Chakrabarty et~al.(2020)Chakrabarty, Muresan, and
  Peng]{chakrabarty2020simile}
T.~Chakrabarty, S.~Muresan, and N.~Peng, ``Generating similes effortlessly like
  a pro: A style transfer approach for simile generation,'' in
  \emph{Proceedings of EMNLP 2020}, 2020.

\bibitem[Kar et~al.(2020)Kar, Aguilar, Lapata, and Solorio]{kar2020multiview}
S.~Kar, G.~Aguilar, M.~Lapata, and T.~Solorio, ``Multi-view story
  characterization from movie plot synopses and reviews,'' in \emph{Proceedings
  of EMNLP 2020}.\hskip 1em plus 0.5em minus 0.4em\relax Punta Cana, Dominican
  Republic: Association for Computational Linguistics, November 2020.

\bibitem[Calderwood et~al.(2020)Calderwood, Qiu, Gero, and
  Chilton]{Calderwood2020HowNU}
A.~Calderwood, V.~Qiu, K.~Gero, and L.~B. Chilton, ``How novelists use
  generative language models: An exploratory user study,'' in
  \emph{HAI-GEN+user2agent@IUI}, 2020.

\bibitem[Osone et~al.(2021)Osone, Lu, and Ochiai]{Osone_2021_BunCho}
\BIBentryALTinterwordspacing
H.~Osone, J.-L. Lu, and Y.~Ochiai, \emph{BunCho: AI Supported Story Co-Creation
  via Unsupervised Multitask Learning to Increase Writers’ Creativity in
  Japanese}.\hskip 1em plus 0.5em minus 0.4em\relax New York, NY, USA:
  Association for Computing Machinery, 2021. [Online]. Available:
  \url{https://doi.org/10.1145/3411763.3450391}
\BIBentrySTDinterwordspacing

\bibitem[Yuan et~al.(2022)Yuan, Coenen, Reif, and
  Ippolito]{Yuan_2022_wordcraft}
\BIBentryALTinterwordspacing
A.~Yuan, A.~Coenen, E.~Reif, and D.~Ippolito, ``Wordcraft: Story writing with
  large language models,'' in \emph{Proceedings of IUI}.\hskip 1em plus 0.5em
  minus 0.4em\relax New York, NY, USA: Association for Computing Machinery,
  2022, p. 841–852. [Online]. Available:
  \url{https://doi.org/10.1145/3490099.3511105}
\BIBentrySTDinterwordspacing

\bibitem[Sander(2013)]{Sander2013}
D.~Sander, ``Models of emotion: the affective neuroscience approach,''
  \emph{The Cambridge Handbook of Human Affective Neuroscience}, pp. 5--53, 01
  2013.

\bibitem[Darwin(1859)]{darwin1859}
C.~Darwin, \emph{On the Origin of Species by Means of Natural Selection}.\hskip
  1em plus 0.5em minus 0.4em\relax London: Murray, 1859, or the Preservation of
  Favored Races in the Struggle for Life.

\bibitem[Darwin(1871)]{darwin1871}
------, \emph{{The descent of man and selection in relation to sex}}.\hskip 1em
  plus 0.5em minus 0.4em\relax London: John Murray, Albemarle Street, 1871.

\bibitem[Darwin(1872)]{darwin1872}
------, \emph{The Expression of the Emotions in Man and Animals}, 1872, the
  original was published 1898 by Appleton, New York. Reprinted 1965 by the
  University of Chicago Press, Chicago and London,.

\bibitem[Tomkins and McCarter(1964)]{Tomkins1964}
\BIBentryALTinterwordspacing
S.~S. Tomkins and R.~McCarter, ``What and where are the primary affects? some
  evidence for a theory,'' \emph{Perceptual and Motor Skills}, vol.~18, no.~1,
  pp. 119--158, 1964, pMID: 14116322. [Online]. Available:
  \url{https://doi.org/10.2466/pms.1964.18.1.119}
\BIBentrySTDinterwordspacing

\bibitem[Barrett(2017)]{Barrett2017}
L.~F. Barrett, \emph{{How emotions are made: The secret life of the
  brain.}}\hskip 1em plus 0.5em minus 0.4em\relax Boston, MA: Houghton Mifflin
  Harcourt, 2017.

\bibitem[Barrett(2020)]{Barrett2020}
L.~Barrett, ``Hypotheses about emotional development in the theory of
  constructed emotion: A response to developmental perspectives on how emotions
  are made,'' \emph{Human Development}, vol.~64, pp. 1--3, 09 2020.

\bibitem[Russell(1994)]{Russell1994IsTU}
J.~A. Russell, ``Is there universal recognition of emotion from facial
  expression? a review of the cross-cultural studies.'' \emph{Psychological
  bulletin}, vol. 115 1, pp. 102--41, 1994.

\bibitem[Barrett(2016)]{Barrett2016_theory_of_constructed_emotion}
\BIBentryALTinterwordspacing
L.~F. Barrett, ``{The theory of constructed emotion: an active inference
  account of interoception and categorization},'' \emph{Social Cognitive and
  Affective Neuroscience}, vol.~12, no.~1, pp. 1--23, 10 2016. [Online].
  Available: \url{https://doi.org/10.1093/scan/nsw154}
\BIBentrySTDinterwordspacing

\bibitem[Diab et~al.(2011)Diab, Yerian, Schauer, Kashyap, Lopez, Hazen, and
  Feldstein]{Diab2011}
D.~L. Diab, L.~Yerian, P.~Schauer, S.~R. Kashyap, R.~Lopez, L.~Hazen, and A.~E.
  Feldstein, ``{Emotional Granularity and Borderline Personality Disorder
  Michael},'' \emph{Journal of Abnormal Psychology}, vol. 120, no.~2, pp.
  414--426, 2011.

\bibitem[Cowen and Keltner(2017)]{Cowen201702247}
\BIBentryALTinterwordspacing
A.~S. Cowen and D.~Keltner, ``Self-report captures 27 distinct categories of
  emotion bridged by continuous gradients,'' \emph{Proceedings of the National
  Academy of Sciences}, 2017. [Online]. Available:
  \url{https://www.pnas.org/content/early/2017/08/30/1702247114}
\BIBentrySTDinterwordspacing

\bibitem[Clayson and Larson(2019)]{CLAYSON201944}
\BIBentryALTinterwordspacing
P.~E. Clayson and M.~J. Larson, ``The impact of recent and concurrent affective
  context on cognitive control: An erp study of performance monitoring,''
  \emph{International Journal of Psychophysiology}, vol. 143, pp. 44 -- 56,
  2019. [Online]. Available:
  \url{http://www.sciencedirect.com/science/article/pii/S0167876019301795}
\BIBentrySTDinterwordspacing

\bibitem[Gendron et~al.(2012)Gendron, Lindquist, Barsalou, and
  Barrett]{Gendron2012}
\BIBentryALTinterwordspacing
M.~Gendron, K.~A. Lindquist, L.~Barsalou, and L.~F. Barrett,
  ``\BIBforeignlanguage{eng}{Emotion words shape emotion percepts},''
  \emph{\BIBforeignlanguage{eng}{Emotion (Washington, D.C.)}}, vol.~12, no.~2,
  pp. 314--325, Apr 2012, 22309717[pmid]. [Online]. Available:
  \url{https://pubmed.ncbi.nlm.nih.gov/22309717}
\BIBentrySTDinterwordspacing

\bibitem[Yik et~al.(2013)Yik, Widen, and
  Russell]{doi:10.1080/02699931.2013.763769}
\BIBentryALTinterwordspacing
M.~Yik, S.~C. Widen, and J.~A. Russell, ``The within-subjects design in the
  study of facial expressions,'' \emph{Cognition and Emotion}, vol.~27, no.~6,
  pp. 1062--1072, 2013, pMID: 23390928. [Online]. Available:
  \url{https://doi.org/10.1080/02699931.2013.763769}
\BIBentrySTDinterwordspacing

\bibitem[Nelson and Russell(2016)]{NELSON201649}
\BIBentryALTinterwordspacing
N.~L. Nelson and J.~A. Russell, ``A facial expression of pax: Assessing
  children’s “recognition” of emotion from faces,'' \emph{Journal of
  Experimental Child Psychology}, vol. 141, pp. 49 -- 64, 2016. [Online].
  Available:
  \url{http://www.sciencedirect.com/science/article/pii/S0022096515001836}
\BIBentrySTDinterwordspacing

\bibitem[Martin et~al.(2018)Martin, Ammanabrolu, Wang, Hancock, Singh,
  Harrison, and Riedl]{Martin_AAAI1817046}
\BIBentryALTinterwordspacing
L.~Martin, P.~Ammanabrolu, X.~Wang, W.~Hancock, S.~Singh, B.~Harrison, and
  M.~Riedl, ``Event representations for automated story generation with deep
  neural nets,'' 2018. [Online]. Available:
  \url{https://aaai.org/ocs/index.php/AAAI/AAAI18/paper/view/17046}
\BIBentrySTDinterwordspacing

\bibitem[Sims et~al.(2019)Sims, Park, and Bamman]{sims-etal-2019-literary}
\BIBentryALTinterwordspacing
M.~Sims, J.~H. Park, and D.~Bamman, ``Literary event detection,'' in
  \emph{Proceedings of ACL 2019}.\hskip 1em plus 0.5em minus 0.4em\relax
  Florence, Italy: Association for Computational Linguistics, Jul. 2019, pp.
  3623--3634. [Online]. Available:
  \url{https://www.aclweb.org/anthology/P19-1353}
\BIBentrySTDinterwordspacing

\bibitem[Tambwekar et~al.(2019)Tambwekar, Dhuliawala, Martin, Mehta, Harrison,
  and Riedl]{Tambwekar_ijcai2019-829}
\BIBentryALTinterwordspacing
P.~Tambwekar, M.~Dhuliawala, L.~J. Martin, A.~Mehta, B.~Harrison, and M.~O.
  Riedl, ``Controllable neural story plot generation via reward shaping,'' in
  \emph{Proceedings of IJCAI}.\hskip 1em plus 0.5em minus 0.4em\relax
  International Joint Conferences on Artificial Intelligence Organization, 7
  2019, pp. 5982--5988. [Online]. Available:
  \url{https://doi.org/10.24963/ijcai.2019/829}
\BIBentrySTDinterwordspacing

\bibitem[Rashkin et~al.(2018{\natexlab{a}})Rashkin, Sap, Allaway, Smith, and
  Choi]{rashkin-etal-2018-event2mind}
\BIBentryALTinterwordspacing
H.~Rashkin, M.~Sap, E.~Allaway, N.~A. Smith, and Y.~Choi, ``{E}vent2{M}ind:
  Commonsense inference on events, intents, and reactions,'' in
  \emph{Proceedings of ACL 2018}.\hskip 1em plus 0.5em minus 0.4em\relax
  Melbourne, Australia: Association for Computational Linguistics, Jul. 2018,
  pp. 463--473. [Online]. Available:
  \url{https://www.aclweb.org/anthology/P18-1043}
\BIBentrySTDinterwordspacing

\bibitem[Mostafazadeh et~al.(2017)Mostafazadeh, Roth, Louis, Chambers, and
  Allen]{mostafazadeh-EtAl:2017:LSDSem}
\BIBentryALTinterwordspacing
N.~Mostafazadeh, M.~Roth, A.~Louis, N.~Chambers, and J.~Allen, ``Lsdsem 2017
  shared task: The story cloze test,'' in \emph{Proceedings of the 2nd Workshop
  on Linking Models of Lexical, Sentential and Discourse-level
  Semantics}.\hskip 1em plus 0.5em minus 0.4em\relax Valencia, Spain:
  Association for Computational Linguistics, April 2017, pp. 46--51. [Online].
  Available: \url{http://aclweb.org/anthology/W17-0906}
\BIBentrySTDinterwordspacing

\bibitem[Rashkin et~al.(2018{\natexlab{b}})Rashkin, Bosselut, Sap, Knight, and
  Choi]{rashkin-etal-2018-modeling}
\BIBentryALTinterwordspacing
H.~Rashkin, A.~Bosselut, M.~Sap, K.~Knight, and Y.~Choi, ``Modeling naive
  psychology of characters in simple commonsense stories,'' in
  \emph{Proceedings of ACL 2018}.\hskip 1em plus 0.5em minus 0.4em\relax
  Melbourne, Australia: Association for Computational Linguistics, Jul. 2018,
  pp. 2289--2299. [Online]. Available:
  \url{https://www.aclweb.org/anthology/P18-1213}
\BIBentrySTDinterwordspacing

\bibitem[Mostafazadeh et~al.(2020)Mostafazadeh, Kalyanpur, Moon, Buchanan,
  Berkowitz, Biran, and Chu-Carroll]{mostafazadeh2020glucose}
N.~Mostafazadeh, A.~Kalyanpur, L.~Moon, D.~Buchanan, L.~Berkowitz, O.~Biran,
  and J.~Chu-Carroll, ``{GLUCOSE}: Generalized and contextualized story
  explanations,'' in \emph{Proceedings of EMNLP 2020}.\hskip 1em plus 0.5em
  minus 0.4em\relax Punta Cana, Dominican Republic: Association for
  Computational Linguistics, November 2020.

\bibitem[Grosz et~al.(1995)Grosz, Joshi, and
  Weinstein]{grosz-etal-1995-centering}
\BIBentryALTinterwordspacing
B.~J. Grosz, A.~K. Joshi, and S.~Weinstein, ``{C}entering: A framework for
  modeling the local coherence of discourse,'' \emph{Computational
  Linguistics}, vol.~21, no.~2, pp. 203--225, 1995. [Online]. Available:
  \url{https://www.aclweb.org/anthology/J95-2003}
\BIBentrySTDinterwordspacing

\bibitem[Barzilay and Lapata(2008)]{Barzilay_2008_local_coherence}
\BIBentryALTinterwordspacing
R.~Barzilay and M.~Lapata, ``Modeling local coherence: An entity-based
  approach,'' \emph{Computational Linguistics}, vol.~34, no.~1, pp. 1--34,
  2008. [Online]. Available: \url{https://doi.org/10.1162/coli.2008.34.1.1}
\BIBentrySTDinterwordspacing

\bibitem[Clark et~al.(2018{\natexlab{b}})Clark, Ji, and
  Smith]{clark-etal-2018-neural}
\BIBentryALTinterwordspacing
E.~Clark, Y.~Ji, and N.~A. Smith, ``Neural text generation in stories using
  entity representations as context,'' in \emph{Proceedings of NAACL-HLT
  2018}.\hskip 1em plus 0.5em minus 0.4em\relax New Orleans, Louisiana:
  Association for Computational Linguistics, Jun. 2018, pp. 2250--2260.
  [Online]. Available: \url{https://www.aclweb.org/anthology/N18-1204}
\BIBentrySTDinterwordspacing

\bibitem[Bamman et~al.(2019)Bamman, Popat, and
  Shen]{bamman-etal-2019-annotated}
\BIBentryALTinterwordspacing
D.~Bamman, S.~Popat, and S.~Shen, ``An annotated dataset of literary
  entities,'' in \emph{Proceedings of NAACL-HLT 2019)}.\hskip 1em plus 0.5em
  minus 0.4em\relax Minneapolis, Minnesota: Association for Computational
  Linguistics, Jun. 2019, pp. 2138--2144. [Online]. Available:
  \url{https://www.aclweb.org/anthology/N19-1220}
\BIBentrySTDinterwordspacing

\bibitem[Bamman et~al.(2020)Bamman, Lewke, and
  Mansoor]{bamman-etal-2020-annotated}
\BIBentryALTinterwordspacing
D.~Bamman, O.~Lewke, and A.~Mansoor, ``\BIBforeignlanguage{English}{An
  annotated dataset of coreference in {E}nglish literature},'' in
  \emph{\BIBforeignlanguage{English}{Proceedings of LREC 2020}}.\hskip 1em plus
  0.5em minus 0.4em\relax Marseille, France: European Language Resources
  Association, May 2020, pp. 44--54. [Online]. Available:
  \url{https://www.aclweb.org/anthology/2020.lrec-1.6}
\BIBentrySTDinterwordspacing

\bibitem[Propp(1968)]{Propp1928}
V.~I. Propp, \emph{Morphology of the Folktale (Translated by L. Scott)}.\hskip
  1em plus 0.5em minus 0.4em\relax University of Texas Press, 1968.

\bibitem[{Imabuchi} and {Ogata}(2012)]{Imabuchi_and_Ogata_conference}
S.~{Imabuchi} and T.~{Ogata}, ``Story generation system based on propp theory
  as a mechanism in narrative generation system,'' in \emph{2012 IEEE Fourth
  International Conference On Digital Game And Intelligent Toy Enhanced
  Learning}, 2012, pp. 165--167.

\bibitem[Peinado and Gerv{\'a}s(2005)]{peinado2005creativity}
F.~Peinado and P.~Gerv{\'a}s, ``Creativity issues in plot generation,'' in
  \emph{Workshop on Computational Creativity, Working Notes, 19th International
  Joint Conference on AI}, 2005, pp. 45--52.

\bibitem[Gerv{\'a}s et~al.(2006)Gerv{\'a}s, L{\"o}nneker-Rodman, Meister, and
  Peinado]{Gervs2006NarrativeM}
P.~Gerv{\'a}s, B.~L{\"o}nneker-Rodman, J.~Meister, and F.~Peinado, ``Narrative
  models : Narratology meets artificial intelligence,'' 2006.

\bibitem[Imabuchi and Ogata(2012)]{Imabuchi_and_Ogata_2012_journal}
S.~Imabuchi and T.~Ogata, ``A story generation system based on propp theory: As
  a mechanism in an integrated narrative generation system,'' in \emph{Advances
  in Natural Language Processing}, H.~Isahara and K.~Kanzaki, Eds.\hskip 1em
  plus 0.5em minus 0.4em\relax Berlin, Heidelberg: Springer Berlin Heidelberg,
  2012, pp. 312--321.

\bibitem[Ogata and Akimoto(2019)]{Ogata2019_book}
\BIBentryALTinterwordspacing
T.~Ogata and T.~Akimoto, \emph{Post-narratology through computational and
  cognitive approaches}, ser. Advances in linguistics and communication studies
  (ALCS) book series.\hskip 1em plus 0.5em minus 0.4em\relax Information
  Science Reference, 2019. [Online]. Available:
  \url{https://ci.nii.ac.jp/ncid/BB29667475}
\BIBentrySTDinterwordspacing

\bibitem[Bosselut et~al.(2019)Bosselut, Rashkin, Sap, Malaviya, Celikyilmaz,
  and Choi]{bosselut-etal-2019-comet}
\BIBentryALTinterwordspacing
A.~Bosselut, H.~Rashkin, M.~Sap, C.~Malaviya, A.~Celikyilmaz, and Y.~Choi,
  ``{COMET}: Commonsense transformers for automatic knowledge graph
  construction,'' in \emph{Proceedings of ACL 2019}.\hskip 1em plus 0.5em minus
  0.4em\relax Florence, Italy: Association for Computational Linguistics, Jul.
  2019, pp. 4762--4779. [Online]. Available:
  \url{https://www.aclweb.org/anthology/P19-1470}
\BIBentrySTDinterwordspacing

\bibitem[Mohammad()]{mohammad-2018-LREC_slide}
S.~Mohammad, ``Word affect intensities,''
  \url{https://www.saifmohammad.com/WebDocs/LREC2018-word-emo-talk.pdf},
  accessed: October 23, 2020.

\end{thebibliography}

\end{document}